\documentclass[lettersize,journal]{IEEEtran}
\usepackage{amsmath,amssymb,amsfonts}%
\usepackage{amsthm}%
\usepackage{mathrsfs}%
\usepackage{algorithm}
\usepackage{algpseudocode}%
\usepackage{array}
\usepackage{textcomp}
\usepackage{stfloats}
\usepackage{url}
\usepackage{verbatim}
\usepackage{graphicx, float}%
\usepackage{multirow}%
\usepackage{cite}
\usepackage{xcolor}%
\usepackage{textcomp}%
\usepackage{manyfoot}%
\usepackage{booktabs}%
\usepackage{listings}%
\usepackage{subcaption} 
\usepackage{dcolumn}
\usepackage[utf8]{inputenc} 
\usepackage{soul}
\usepackage{wrapfig}
\usepackage{soul,color}
\usepackage{makecell}
\usepackage{comment}
\usepackage{url}
\usepackage{hyperref}
\usepackage{xurl}   
\usepackage[table]{xcolor}
\newcommand{\cmark}{\checkmark} 
\newcommand{\xmark}{X}

\newtheorem{theorem}{Theorem}

\begin{document}

\title{Robust Multispectral Semantic Segmentation under Missing or Full Modalities via Structured Latent Projection}

\author{Irem~Ulku, ~Erdem~Akagündüz, 
and Ömer~Özgür~Tanrıöver
\thanks{I. Ulku (corresponding author) and O. Tanriover are with the Department
of Computer Engineering, Ankara University, Ankara,
Türkiye. e-mail: {irem.ulku,tanriover}@ankara.edu.tr.}
\thanks{E. Akagunduz is with the Department of Modeling and Simulation, Graduate School of Informatics, METU, Ankara, Türkiye. email: akaerdem@metu.edu.tr}
}

\markboth{}
{Shell \MakeLowercase{\textit{et al.}}: A Sample Article Using IEEEtran.cls for IEEE Journals}


\maketitle

\begin{abstract}
Multimodal remote sensing data provide complementary information for semantic segmentation, but in real-world deployments, some modalities may be unavailable due to sensor failures, acquisition issues, or challenging atmospheric conditions. Existing multimodal segmentation models typically address missing modalities by learning a shared representation across inputs. However, this approach can introduce a trade-off by compromising modality-specific complementary information and reducing performance when all modalities are available. In this paper, we tackle this limitation with CBC-SLP, a multimodal semantic segmentation model designed to preserve both modality-invariant and modality-specific information. Inspired by the theoretical results on modality alignment, which state that perfectly aligned multimodal representations can lead to sub-optimal performance in downstream prediction tasks, we propose a novel structured latent projection approach as an architectural inductive bias. Rather than enforcing this strategy through a loss term, we incorporate it directly into the architecture. In particular, to use the complementary information effectively while maintaining robustness under random modality dropout, we structure the latent representations into shared and modality-specific components and adaptively transfer them to the decoder according to the random modality availability mask. Extensive experiments on three multimodal remote sensing image sets demonstrate that CBC-SLP consistently outperforms state-of-the-art multimodal models across full and missing modality scenarios. Besides, we empirically demonstrate that the proposed strategy can recover the complementary information that may not be preserved in a shared representation. The code is available at \url{https://github.com/iremulku/Multispectral-Semantic-Segmentation-via-Structured-Latent-Projection-CBC-SLP-}.
\end{abstract}


\section{Introduction}
\begin{figure}[t]
\centering
\includegraphics[width=0.24\textwidth]{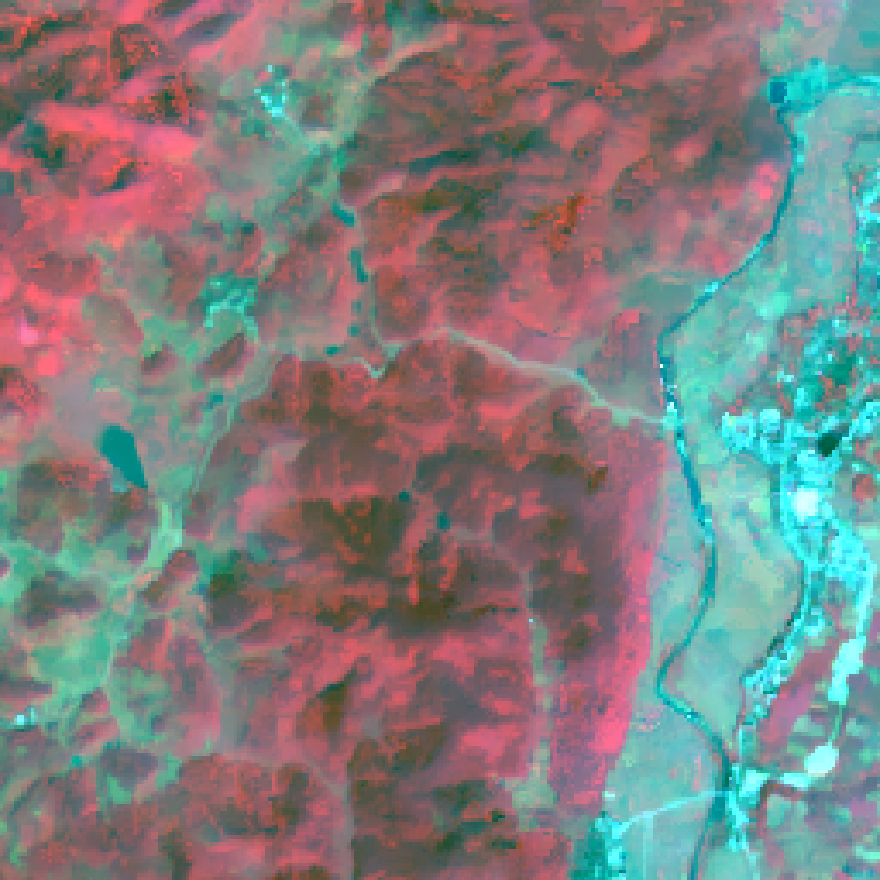}
\includegraphics[width=0.24\textwidth]{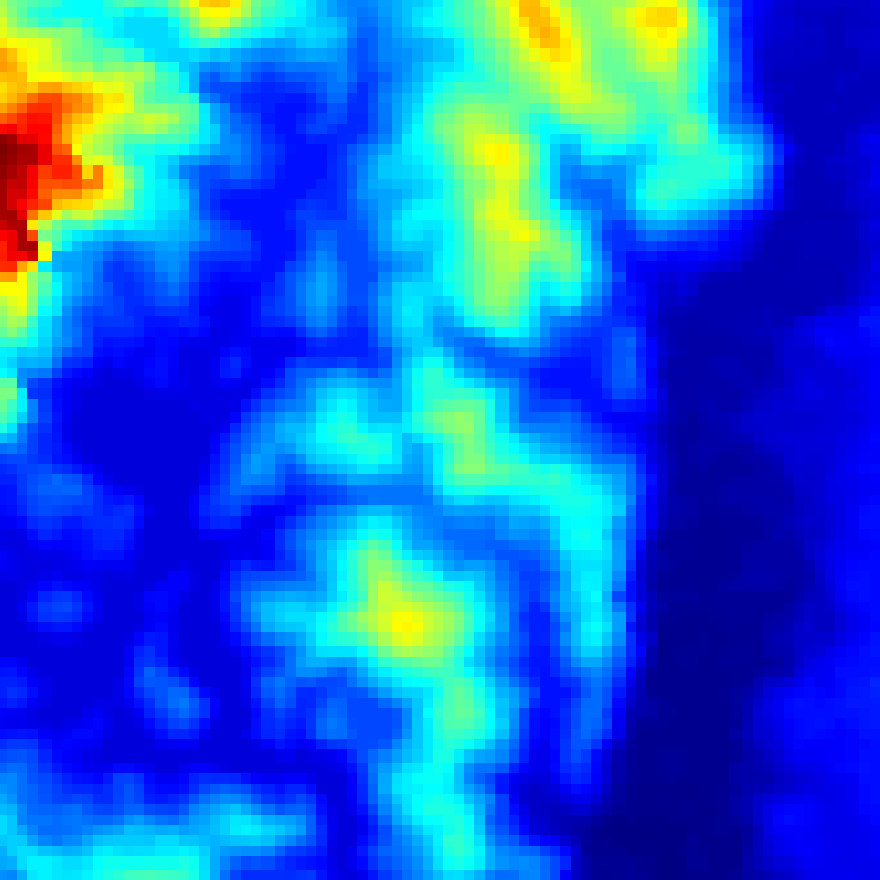}
\caption{Multispectral optical (false RGB: infrared-red-green channels) image on the left and DEM height map on the right, both from the Hunan image set \cite{li2022dkdfn}. While the multispectral image highlights surface reflectance and land-cover differences, the DEM height map reveals the terrain structure, providing complementary information for scene interpretation.}\label{ex}
\end{figure}

\IEEEPARstart{A}{ccurate} land-cover mapping is vital for Earth observation applications. In particular, remote sensing semantic segmentation \cite{wang2026gift, chen2025rest}, which enables large-scale mapping of crops, trees, and forests, supports agricultural monitoring \cite{zaheer2023season} and ecological analysis \cite{liu2024clusterformer}. However, relying on a single remote sensing modality limits segmentation performance, since complementary modalities provide mutually reinforcing information for class separation \cite{stathaki2008image, liu2024infrared}. Multispectral optical data, including RGB, near-infrared (NIR), and short-wave infrared (SWIR) bands, enhance class discrimination by providing spectral cues sensitive to chlorophyll and moisture \cite{ulku2022deep}. Height and terrain information from digital surface models (DSM) and digital elevation models (DEM) improves separability by providing geometric context that complements spectral information \cite{zhou2023dsm}. To illustrate the benefit of multimodal fusion, Fig. \ref{ex} shows an example from Hunan, China, where the Sentinel-2 multispectral optical image is paired with a DEM height map over the same area to highlight complementary spectral and topographic cues for land-cover mapping  \cite{li2022dkdfn}. Moreover, synthetic aperture radar (SAR) offers structure-sensitive information that enhances robustness, particularly when optical imagery is degraded under challenging illumination or atmospheric conditions \cite{wu2025segcr}. Therefore, integrating these complementary modalities is promising for achieving accurate semantic segmentation across diverse scenes \cite{han2025multimodal, xu2023multimodal}.

Although fusing remote sensing data with different spatial, spectral, and angular resolutions is effective \cite{gomez2015multimodal}, some modalities are not always available, e.g., due to {varying electro-optical configurations,} malfunctions in satellite sensors or challenging atmospheric conditions\cite{ zhang2018missing, wei2023msh, dalla2015challenges}. For instance, a hardware failure in the onboard imaging instrument of the Landsat-7 satellite platform {may} lead to systematic data gaps in the images \cite{liu2023decrecnet}. Similarly, due to the dense cloud cover during the acquisition of MODIS sensor observations, the resulting satellite images exhibit information loss \cite{yu2024missing}. SAR data collection is also challenging because certain surface conditions reduce coherence, and processing errors often yield low raw data quality \cite{hippert2020eof}. These challenges lead to the failure of existing fusion models, as these approaches generally rely on the complete availability of all modalities during the training and inference phases \cite{zhou2026remote, chen2024novel}. While current approaches often work well on the modality sets used during training, their performance declines on untrained modalities at test time \cite{reza2024robust}. Although training separate networks for each possible modality combination is conceptually straightforward, this strategy becomes {computationally} ineffective when the number of input modalities is large.

Recent multimodal semantic segmentation approaches for remote sensing under incomplete modality settings aggregate complementary cross-modal information and compensate for the missing modality through transfer, reconstruction, or relation propagation. For instance, MetaRS \cite{zhou2026remote} jointly learns modality-specific representations and a shared meta-modal representation across all modalities in the EarthMiss benchmark. Then, the missing information is recovered via transfer in this meta-modal space. RobSense proposes a single foundation model in which missing inputs are restored by a unimodal latent reconstructor when spectral bands or temporal observations are absent \cite{do2025robsense}. The multimodal transformer in \cite{chen2024novel} integrates heterogeneous modalities into a unified representation using learnable fusion tokens and modality attention. This approach is capable of handling missing modalities both under training and inference phases via masked self-attention. MHHL \cite{han2025multimodal} models cross-modal relations through a heterogeneous hypergraph and employs information propagation to compensate for missing modalities. Primarily, these approaches focus on improving performance under missing-modality scenarios and rarely address the trade-off between missing modality robustness and optimal accuracy when all modalities are available. However, {we posit that} the nature of the challenge lies in achieving a unified fusion mechanism that is robust to missing modalities without sacrificing performance when all modalities are available. Given this observation, we aim to develop a unified semantic segmentation model that operates effectively in both cases.

Empirical results have shown that exact modality alignment is generally sub-optimal for downstream prediction tasks; instead, constructing meaningful latent modality structures that preserve both shared and modality-specific information can improve model performance \cite{jiang2023understanding}. Our key insight is that by projecting a multimodal representation into shared and modality-specific subspaces, the model may retain useful information under missing modality conditions while also improving representational capacity when all modalities are available. Specifically, we introduce a novel structure 
that splits the {latent} multimodal representation into shared and modality-specific components, improving robustness under random modality dropout while mitigating the performance drop in the full-modality setting. 

Recently, \cite{li2025simmlm} introduced a framework that targets strong performance under both full and missing modality settings by explicitly modeling missing modality conditions through additional loss terms. In contrast to such loss-based designs, we {split} and recombine the shared and modality-specific representations at the architectural level throughout the pipeline, rather than relying on extra loss terms. Since explicit loss-based alignment does not always guarantee meaningful latent modality structures \cite{zhao2025freefusion}, we aim to introduce inductive bias through structured latent space projection without relying on additional adaptations of loss functions. Therefore, instead of compressing correlation-aware features into a single latent space as in \cite{ulku2025cross}, we adopt a two-part decomposition that isolates shared correlation and preserves modality-specific information. Projecting the latent representation into shared and modality-specific components reduces gradient interference between full and missing conditions \cite{wei2025boosting}. In this way, the decoder {sees} a consistent input structure across different modality availability patterns. 

Overall, the contributions of this work are listed as follows:

\begin{itemize}
    \item We propose an improved version of our prior model \cite{ulku2025cross}, namely CBC-SLP, which maintains good performance in both full and missing modality scenarios by imposing an architectural inductive bias through shared and modality-specific latent decompositions.
    \item We demonstrate that the proposed CBC-SLP model is not only beneficial for missing modalities but also performs well in a full modality scenario. Additionally, we {show} the generalization capability of this model on both homogeneous and heterogeneous modality configurations.
    \item The extensive experiments and analyses on three multimodal remote sensing image sets across multiple land-cover categories demonstrate that our approach outperforms others for missing modality cases while reducing the performance drop in the full modality setting.
\end{itemize}

\section{Related Works}
\subsection{Semantic Segmentation with Missing Modalities}

Semantic segmentation approaches for the missing modality problem can be categorized into two main groups. The first group of studies utilize data generation with generative architectures and synthesize the missing modalities based on the available input modalities \cite{yu2019ea}. {The primary weakness of these approaches is that segmentation accuracy is largely dependent on the quality of the synthesized modalities.} 

The second approach involves learning a shared latent representation that captures common features across the available modalities, leading to more reliable segmentation. Approaches for handling missing modality cases can be stated in four subcategories. For example, RFNet \cite{ding2021rfnet} directly fuses features from the available modalities via region-aware attention and enhances robustness to missing modalities through a regularizer. However, it is not trivial to directly fuse the available modalities due to the combinatorial diversity of possible modality subsets. 

A different approach for learning a shared latent representation is to extract modality-invariant and modality-specific content codes via feature disentanglement. This strategy is explored in \cite{liu2025kmd}, where the authors leverage a Koopman Invariant Subspace to disentangle modality-common and modality-specific information and predict modality relationships under missing-modality settings to reconstruct the original input modalities.

Another option for learning a shared latent representation is to adapt a student network to a teacher network via domain adaptation. Representative works in this category are based on either parameter-efficient adaptation \cite{reza2024robust} that modulates intermediate features with few additional parameters or designing a teacher–student model that employs margin-aware distillation and modality-aware regularization for incomplete multimodal learning \cite{wei2023mmanet}. Margin-aware distillation encourages the network to focus on samples close to the decision boundary, while the modality-aware regularization adaptively enhances representational capacity for weaker modality combinations.

Final subcategory methods use the correlations across input modalities to learn a shared latent representation. These methods extract the correlation representation for each modality that can recover the missing modality features. One such work is the Learnable Sorting State Space Model
(LS3M) \cite{zhang2025incomplete}, where differentiable permutation matrices reorder the input sequences of available modalities based on modality-specific characteristics. This dynamic reordering preserves long-range semantic correlations that are important for incomplete modalities. While MMMViT \cite{qiu2024mmmvit} models cross-modal correlations to explicitly recover missing-modality features, its design is primarily focused on robustness across arbitrary modality subsets.

\subsection{Semantic Segmentation With Missing Modalities for Remote Sensing}

Few multimodal models have been proposed to address the challenges of missing remote sensing modalities. Two predominant semantic segmentation approaches for this problem fall into either missing modality synthesis or shared latent representation learning. Synthesis approaches aim to generate the missing remote sensing modality from the available modalities acquired over the same area. RobSense \cite{do2025robsense} proposes a robust multimodal foundation model for multispectral and SAR data. It uses two unimodal latent reconstructors to recover rich representations from incomplete inputs with missing spectral bands or irregular temporal sequences. However, the drawback is that incompleteness is simulated by randomly masking spectral bands or time points, so robustness to real-world missing patterns cannot be thoroughly assessed. 

Shared latent representation learning approaches usually learn a unified embedding from the available modalities to perform segmentation under missing remote sensing inputs. One strategy is to directly fuse information from the available modalities into a single unified representation, in which the model can still operate when some inputs are missing. For example, \cite{chen2024novel} introduces learnable fusion tokens and uses modality attention together with masked self-attention to collect multimodal signals, while allowing random modality combinations during training. Unfortunately, combining mask-reconstruction and contrastive pretraining can yield task-dependent gains and may even underperform simpler training settings.

A shared latent representation can also be learned by disentangling shared content from modality-specific variations. A recent study \cite{zhou2026remote} introduces the EarthMiss benchmark and proposes MetaRS, which disentangles a shared meta-modal representation from modality-specific features and transfers knowledge in the meta-modal space to handle missing modality scenarios in optical and SAR land-cover mapping. MetaRS may underperform when all modalities are fully available, as it is primarily designed for missing modality settings.

Complementary cues often exist across remote sensing modalities, such as spectral, elevation, and radar backscatter patterns that characterize the same land surface. To better exploit this complementarity when some inputs are missing, correlation-modeling approaches estimate cross-modal correlations before fusion and use them to guide the shared representation for segmentation. Relevant to this approach, MHHL \cite{han2025multimodal} addresses incomplete modality semantic segmentation by constructing a multimodal heterogeneous hypergraph and propagating information across different modality availability patterns. 

Aside from this hypergraph-based propagation, we revisit incomplete modality learning from a latent decomposition perspective, where the representation is split into a shared component and modality-specific components and then recombined before decoding.  We first compute cross-band correlation features, preserve modality-specific cues via separate branches, and pass a structured representation to the decoder with missing modalities masked. Unlike the approach in \cite{li2025simmlm}, {where the authors introduce an additional loss regularization term to improve accuracy in both full and partial modality settings, we propose an efficient architecture to achieve this goal.} To the best of our knowledge, achieving a single unified fusion mechanism that is reliable under missing modalities while maintaining full-modality performance remains an open challenge.

\section{Methodology}
In this section, we first describe three benchmark multimodal image sets that are used in our experiments. Following that, we explain our CBC-SLP model, which includes a structured latent projection for robust performance in missing and full modality cases. 

\subsection{Image Sets}
 {We utilize three multimodal remote sensing image sets. Two of these sets (Potsdam and Hunan) have heterogeneous modalities, while the other (DSTL) provides homogeneous modalities. For each set}, all images are cropped into 224 × 224 patches. To normalize the inputs, per-channel mean subtraction is applied independently for each modality {seperately}, where the mean values are computed only over the training subset indices and then subtracted from both training and evaluation samples. For DSTL and Potsdam, the images are split into 72\% training, 8\% validation, and 20\% testing sets. For Hunan, we adopt the official split, which assigns 80\% of the tiles to training, 10\% to validation, and 10\% to testing. All tasks are formulated as binary semantic segmentation for target land-cover classes. 

\begin{figure*}[t]
\centering

\begin{subfigure}[t]{0.32\textwidth}
  \centering
  \includegraphics[width=\linewidth]{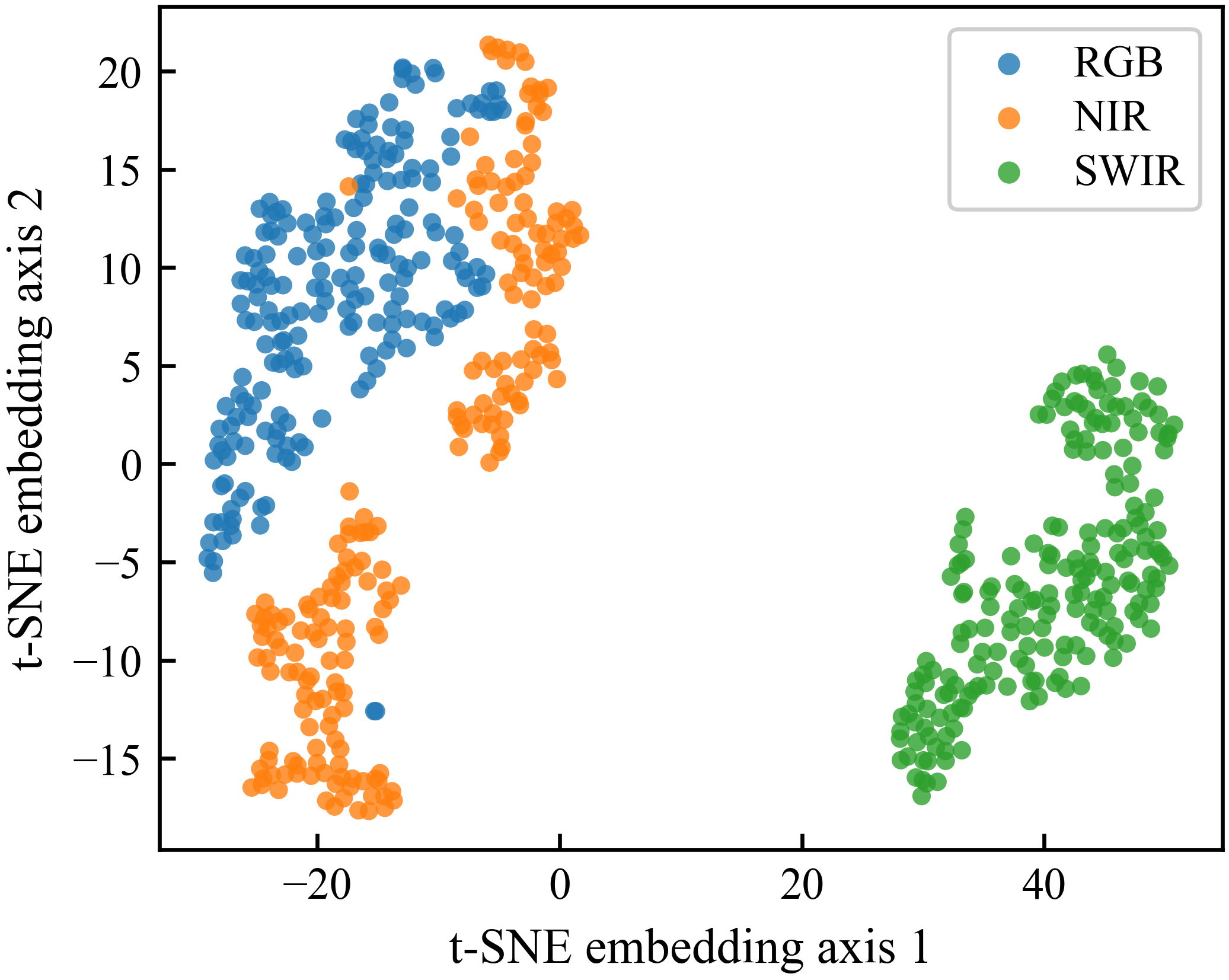}
  \caption{DSTL}
  \label{fig:corr_dstl}
\end{subfigure}\hfill
\begin{subfigure}[t]{0.32\textwidth}
  \centering
  \includegraphics[width=\linewidth]{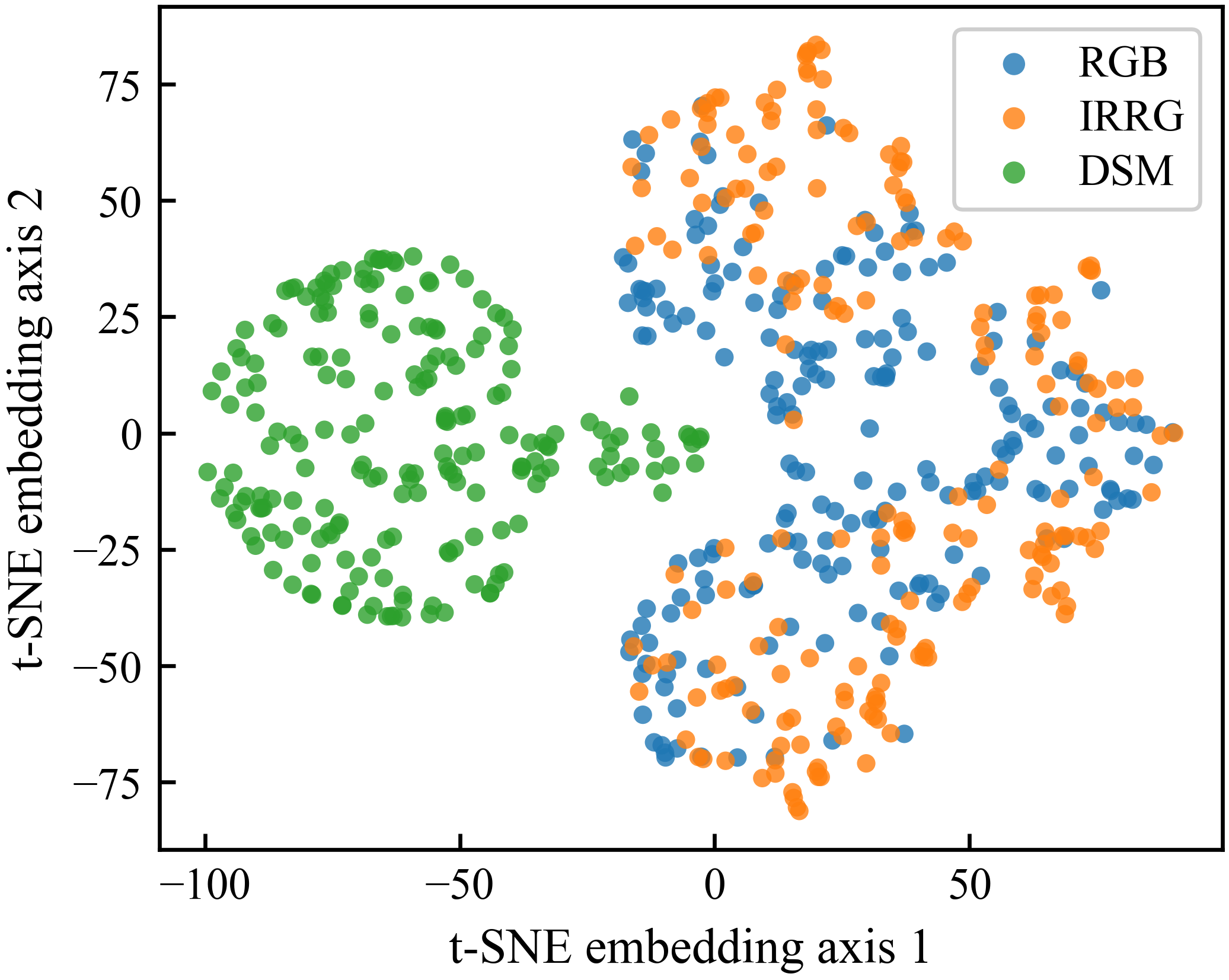}
  \caption{Potsdam}
  \label{fig:corr_potsdam}
\end{subfigure}\hfill
\begin{subfigure}[t]{0.32\textwidth}
  \centering
  \includegraphics[width=\linewidth]{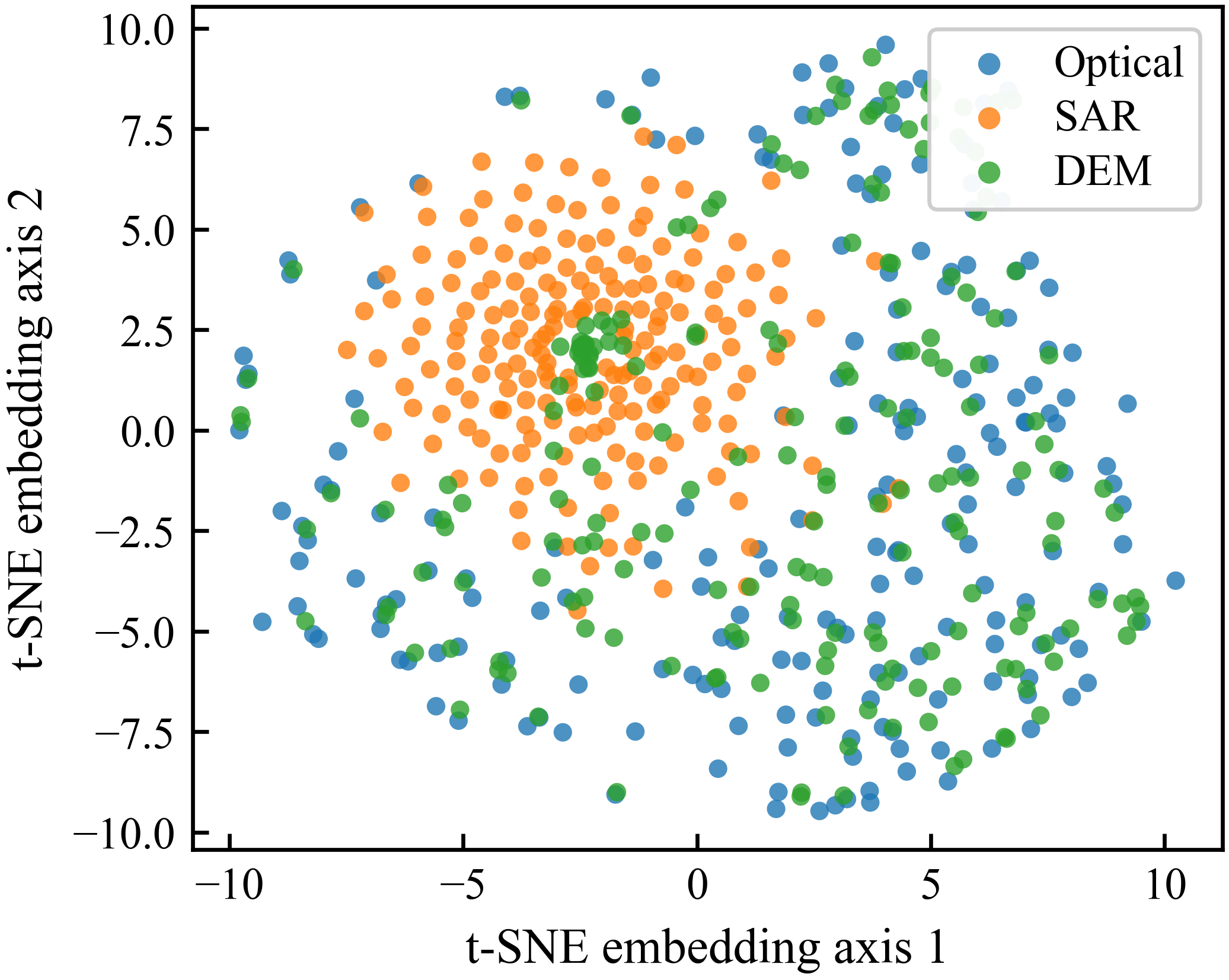}
  \caption{Hunan}
  \label{fig:corr_hunan}
\end{subfigure}

\captionsetup{font=normalsize} 
\captionsetup[sub]{font=normalsize} 
\caption{Cross-modality {T-distributed stochastic neighbor embedding} distributions on DSTL, Potsdam, and Hunan {image sets}.}
\label{fig:modality_corr_all}
\end{figure*}

{In order to illustrate the cross-modal relations for each set} Fig.~\ref{fig:modality_corr_all} {presents cross-modality T-distributed stochastic neighbor embeddings (t-SNE) of for each modality,} where cluster overlap indicates shared information and separation indicates discriminative information. The least correlated modalities are SWIR, SAR, and DSM for DSTL, Hunan, and Potsdam, respectively, indicating complementary information. {Specific details for each dataset is provided in the following.}

        \label{fig:corr_dstl}


\paragraph{{DSTL Satellite Imagery Feature Detection Image Set}}

The DSTL satellite imagery feature detection image set is released by the UK Defence Science and Technology Laboratory for the Kaggle competition \cite{dstl_kaggle}. DSTL contains 25 high-resolution satellite image tiles collected by WorldView-3 covering a 1 km × 1 km area. Each scene has visible RGB bands in the 450–690 nm range, NIR bands in the 770–1040 nm range, and the SWIR coverage spans eight bands ranging from 1195-2365 nm. The WorldView-3 optical system delivers imagery at different spatial resolutions, with about 0.31 m in the visible bands, 1.24 m in the multispectral bands, and roughly 7.5 m in the SWIR bands. 


Visible RGB images generally capture appearance due to color pigment absorption. On the other hand, NIR images have an important role in discriminating leaf internal structure and chlorophyll content, while the SWIR images are sensitive to water content and certain minerals. Therefore, in {our} experiments, DSTL is organized into a trimodal image set using RGB, NIR, and SWIR modalities. Only the crop class is selected for DSTL because other annotations are imbalanced, and crops are the most frequently labeled class.



{However, in real-world environments, many factors such as atmospheric conditions and sensor degradations may lead to partial or even total lack of spectral bands} \cite{neagoe2023band}. Using complementary spectral information may help to cope with the missing data in spectral bands.


\paragraph{{ISPRS 2D Semantic Labeling Potsdam Image Set}}

The Potsdam benchmark from the International Society for Photogrammetry and Remote Sensing (ISPRS) \cite{isprs_potsdam} is used as a trimodal image set consisting of RGB, IRRG (NIR-R-G), and {DSM}. This image set comprises 38 high-resolution aerial image tiles, each measuring 6000 × 6000 pixels, with a 0.05 m ground sampling distance for all modalities. As heterogeneous modalities, RGB, IRRG, and the DSM are used, where the binary semantic segmentation is conducted for the tree land-cover class. 


Different from DSTL, height and terrain information in the DSM modality provides geometric descriptions of the topographic surface of the land coverage. {The generated DSM, however, may not be available due to the noise caused by temporal changes or matching errors} \cite{bittner2019dsm}. 


\paragraph{{Hunan Image Set}}

The third {set, namely Hunan image set} \cite{li2022dkdfn} combines Sentinel-2 optical imagery, Sentinel-1 SAR data, and DEM to provide complementary spectral, structural, and topographic cues. Sentinel-2 multispectral observations in the 492.7–2202.4 nm wavelength range are used, covering the RGB, NIR, and SWIR bands. The Sentinel-1 SAR modality is used in dual polarization. DEM modality is composed of elevation and slope layers provided by the NASA Shuttle Radar Topographic Mission (SRTM) \cite{cgiar_srtm}.

The three modalities have different native spatial resolutions by 10–20 m for Sentinel-2 multispectral images, 10 m for Sentinel-1 SAR, and 30 m for DEM. Therefore, all data are resampled to the finest resolution of 10 m by using the nearest neighbor strategy. The Hunan image set comprises 500 non-overlapping tiles, corresponding to a total of 32,768,000 labeled pixels. A highly dominating forest land-cover class is used in the binary semantic segmentation task, which covers a much greater proportion of the land than other classes. 



\subsection{Proposed Model}
An overview of the proposed CBC-SLP architecture is shown in Fig.~\ref{fig:latent}. Consider $M$ modalities indexed by $m\in\mathcal{M}=\{1,\ldots,M\}$.
For a sample, the multimodal input is denoted by $\mathbf{X}=[\mathbf{X}_1,\mathbf{X}_2,\ldots,\mathbf{X}_M]$,
where this tensor satisfies $\mathbf{X} \in \mathbb{R}^{B\times U\times D\times H\times W}$. Here, $B$ represents batch, $U$ denotes modality, $D$ is the channel, $H$ and $W$ are the pixel dimensions.

A binary availability mask is introduced to model missing modalities, denoted by $\mathbf{s}\in\{0,1\}^{B\times M}$, where $s_{b,m}=0$ indicates that modality $m$ is missing for sample $b$.
The mask is applied at both the encoder level and the latent level through elementwise multiplicative gating.

\subsection{Modality-specific Encoders}
Each modality is processed by a dedicated encoder $E_m(\cdot)$, namely intra-modal encoders (see Figure \ref{fig:latent}). In this study, $E_m(\cdot)$ is implemented as a ResNet-based three-dimensional convolutional encoder obtained by inflating a ResNet50 backbone. Given the modality input tensor $\mathbf{X}_m$, the encoder produces feature maps as follows:
\begin{equation}
\big(\mathbf{x}^{(1)}_{m},\mathbf{x}^{(2)}_{m},\mathbf{x}^{(3)}_{m},\mathbf{x}^{(4)}_{m},\mathbf{x}^{(5)}_{m},\mathbf{x}^{(6)}_{m}\big)
=
E_m(\mathbf{X}_m),
\end{equation}
where, $\mathbf{x}^{(l)}_{m}$ denotes the feature map extracted at scale level $l$.

To explicitly account for missing modalities, gating is applied at every encoder level. Let $s_{b,m}$ denote the availability indicator of modality $m$ across the batch. The gated feature maps are computed by elementwise multiplication as follows:
\begin{equation}
\tilde{\mathbf{x}}^{(l)}_{m}
=
s_{b,m}\odot \mathbf{x}^{(l)}_{m},
\qquad
l\in\{1,\ldots,6\}.
\end{equation}

\subsection{Cross-modal Fusion}
The fusion module follows the cross-band fusion strategy introduced in our previous work \cite{ulku2025cross} and extends it to a general cross-modal setting. Fusion is performed at each encoder level by concatenating the gated modality features along the channel dimension and projecting the result with a $1\times1\times1$ convolution as shown in Fig.~\ref{fig:latent}. For level $l$, the fused representation is formulated as follows:

\begin{equation}
\begin{aligned}
\mathbf{f}^{(l)}
&=
\Phi^{(l)}\!\Big(
\tilde{\mathbf{x}}^{(l)}_{1},
\tilde{\mathbf{x}}^{(l)}_{2},
\ldots,
\tilde{\mathbf{x}}^{(l)}_{M}
\Big),
\qquad
l\in\{1,\ldots,6\}.
\end{aligned}
\end{equation}

The operator $\Phi^{(l)}(\cdot)$ denotes channel-wise concatenation followed by a $1\times1\times1$ convolution, an activation function, and a normalization layer. The fused features $\mathbf{f}^{(1)},\ldots,\mathbf{f}^{(4)}$ are forwarded to the decoder through skip connections, while $\mathbf{f}^{(6)}$ is used to construct the Transformer-based latent representation.

\subsection{Intra-modal Self-Attention Block}
For each modality $m\in\mathcal{M}$, the deepest gated encoder feature map $\tilde{\mathbf{x}}^{(6)}_{m}$ is projected to the Transformer embedding dimension $C_t$ using a $1\times1\times1$ convolution, yielding $\mathbf{F}^{\mathrm{proj}}_{m}\in\mathbb{R}^{B\times C_t\times D_t\times H_t\times W_t}$ as follows:

\begin{equation}
\mathbf{F}^{\mathrm{proj}}_{m}
=
\mathrm{Conv}^{1\times1\times1}_{m}\!\left(\tilde{\mathbf{x}}^{(6)}_{m}\right),
\end{equation}

Tokenization converts $\mathbf{F}^{\mathrm{proj}}_{m}$ into a sequence of $N$ tokens by partitioning the feature map into non-overlapping 3D patches of size $P\times P\times P$, flattening each patch, and using a learnable projection. The resulting token sequence is written as $\mathbf{F}^{\mathrm{token}}_{m}\in\mathbb{R}^{B\times N\times C_t}$, where $N=D_tH_tW_t/P^3$ is the number of image patches. A learnable positional embedding $\mathbf{E}^{\mathrm{pos}}_{m}\in\mathbb{R}^{N\times C_t}$ is added to preserve spatial ordering. The flattened patch features are represented by $\mathbf{F}^{\mathrm{patch}}_{m}\in\mathbb{R}^{B\times N\times P^3C_t}$, and the learnable token projection is given by $\mathbf{E}_{m}\in\mathbb{R}^{P^3C_t\times C_t}$. The token sequence is then obtained as follows:
\begin{equation}
\mathbf{F}^{\mathrm{token}}_{m}
=
\mathbf{F}^{\mathrm{patch}}_{m}\mathbf{E}_{m}
+
\mathbf{E}^{\mathrm{pos}}_{m}.
\end{equation}

The intra-modal Transformer applies multi-head self-attention to $\mathbf{F}^{\mathrm{token}}_{m}$.
Let $L$ denote the number of attention heads and let $d_k=C_t/L$.
For each head $i\in\{1,\ldots,L\}$, the projection tensors satisfy
$\mathbf{W}^{Q,i}_{m}\in\mathbb{R}^{C_t\times d_k}$,
$\mathbf{W}^{K,i}_{m}\in\mathbb{R}^{C_t\times d_k}$,
$\mathbf{W}^{V,i}_{m}\in\mathbb{R}^{C_t\times d_k}$,
and the output projection satisfies $\mathbf{W}^{O}_{m}\in\mathbb{R}^{Ld_k\times C_t}$.
The queries, keys, and values are computed as follows:

\begin{equation}
\begin{aligned}
\mathbf{Q}^{i}_{m}
&=
\mathrm{LN}\!\left(\mathbf{F}^{\mathrm{token}}_{m}\right)\mathbf{W}^{Q,i}_{m},\\
\mathbf{K}^{i}_{m}
&=
\mathrm{LN}\!\left(\mathbf{F}^{\mathrm{token}}_{m}\right)\mathbf{W}^{K,i}_{m},\\
\mathbf{V}^{i}_{m}
&=
\mathrm{LN}\!\left(\mathbf{F}^{\mathrm{token}}_{m}\right)\mathbf{W}^{V,i}_{m},
\end{aligned}
\end{equation}
where $\mathrm{LN}(\cdot)$ represents layer normalization. Scaled dot-product attention is then given by
\begin{equation}
\mathrm{head}^{i}_{m}
=
\mathrm{softmax}\!\left(
\frac{\mathbf{Q}^{i}_{m}\mathbf{K}^{i\top}_{m}}{\sqrt{d_k}}
\right)\mathbf{V}^{i}_{m}.
\end{equation}

The multi-head self-attention output is obtained via concatenation followed by a linear projection:
\begin{equation}
\mathrm{MSA}_{m}
=
\Big[\mathrm{head}^{1}_{m},\ldots,\mathrm{head}^{L}_{m}\Big]\mathbf{W}^{O}_{m},
\end{equation}
where $\big[\cdot\big]$ denotes concatenation. A Transformer layer updates the token sequence using residual learning and a feed-forward network as follows:

\begin{equation}
\begin{aligned}
\mathbf{U}_{m}
&=
\mathbf{F}^{\mathrm{token}}_{m}
+
\mathrm{MSA}_{m},\\
\mathbf{Z}_{m}
&=
\mathbf{U}_{m}
+
\mathrm{MLP}\!\left(\mathrm{LN}\!\left(\mathbf{U}_{m}\right)\right).
\end{aligned}
\end{equation}

\subsection{Inter-modal Correlation Block}
Inter-modal correlations are modeled to enrich each modality representation with information aggregated from the remaining modalities in a pixel-wise manner. For modality $m\in\mathcal{M}$, the intra-modal Transformer output $\mathbf{Z}_{m}\in\mathbb{R}^{B\times N\times C_t}$ is first reshaped back to a volumetric feature map $\hat{\mathbf{Z}}_{m}\in\mathbb{R}^{B\times C_t\times D_t\times H_t\times W_t}$. Query, key, and value tensors are then obtained by a $1\times1\times1$ convolutional projection as:
\begin{equation}
\left[\mathbf{Q}_{m},\mathbf{K}_{m},\mathbf{V}_{m}\right]
=
\mathrm{Conv}^{1\times1\times1}_{qkv,m}\!\left(\hat{\mathbf{Z}}_{m}\right),
\end{equation}
where $\mathbf{Q}_{m}$, $\mathbf{K}_{m}$, and $\mathbf{V}_{m}$ are elements of $\mathbb{R}^{B\times C_t\times D_t\times H_t\times W_t}$.

As indicated in Fig.~\ref{fig:latent}, for a target modality $m$, inter-modal correlation scores $\mathbf{S}_{m\leftarrow j}$ are computed against every source modality $j\in\mathcal{M}$ by elementwise multiplication ($\odot$) as follows:
\begin{equation}
\mathbf{S}_{m\leftarrow j}
=
\mathbf{Q}_{m}\odot \mathbf{K}_{j},
\qquad
j\in\mathcal{M}.
\end{equation}

Pixel-wise attention weights $\boldsymbol{\alpha}_{m\leftarrow j}$ are obtained by applying a softmax over the modality index $j$ at each batch, channel, and spatial location:
\begin{equation}
\boldsymbol{\alpha}_{m\leftarrow j}
=
\frac{
\exp\!\left(\mathbf{S}_{m\leftarrow j}/\sqrt{|\mathcal{M}|}\right)
}{
\sum\limits_{j'\in\mathcal{M}}
\exp\!\left(\mathbf{S}_{m\leftarrow j'}/\sqrt{|\mathcal{M}|}\right)
}.
\end{equation}
where the denominator means the sum of the scores from every possible source $j'$ into the same target $m$.

The correlation enriched representation for modality $m$ is formed as a weighted sum of value tensors with pixel-wise weights:
\begin{equation}
\mathbf{Z}^{\mathrm{corr}}_{m}
=
\sum\limits_{j\in\mathcal{M}}
\boldsymbol{\alpha}_{m\leftarrow j}\odot \mathbf{V}_{j}.
\end{equation}

To explicitly account for missing modalities, $\mathbf{Z}^{\mathrm{corr}}_{m}$ is gated by the availability indicator $s_{b,m}$ through elementwise multiplication to obtain $\tilde{\mathbf{Z}}^{\mathrm{corr}}_{m}$ as follows:

\begin{equation}
\tilde{\mathbf{Z}}^{\mathrm{corr}}_{m}
=
s_{b,m}\odot \mathbf{Z}^{\mathrm{corr}}_{m},
\qquad.
\end{equation}

The gated correlation feature volume $\tilde{\mathbf{Z}}^{\mathrm{corr}}_{m}$ is reshaped to a token sequence and fused with the intra-modal token stream by a residual connection:
\begin{equation}
\mathbf{Z}^{\mathrm{fused}}_{m}
=
\mathbf{Z}_{m}
+
\mathrm{Reshape}\!\left(\tilde{\mathbf{Z}}^{\mathrm{corr}}_{m}\right),
\end{equation}
where $\mathbf{Z}_{m}$ denotes the intra-modal token sequence before inter-modal enrichment, and $\mathrm{Reshape}(\cdot)$ flattens the spatial grid into the token dimension.

A joint token stream is constructed by concatenating the fused tokens from all modalities:

\begin{equation}
\mathbf{Z}_{\mathrm{multi}}
=
\Big[
\mathbf{Z}^{\mathrm{fused}}_{1},
\mathbf{Z}^{\mathrm{fused}}_{2},
\ldots,
\mathbf{Z}^{\mathrm{fused}}_{M}
\Big].
\end{equation}

The cross-modal fused deep feature $\mathbf{f}^{(6)}$ is projected to $C_t$ by a $1\times1\times1$ convolution, reshaped into tokens, and concatenated with $\mathbf{Z}_{\mathrm{multi}}$ to form the multimodal Transformer input:
\begin{equation}
\mathbf{U}
=
\Big[
\mathbf{Z}_{\mathrm{multi}},
\mathrm{Reshape}\!\left(
\mathrm{Conv}^{1\times1\times1}_{\mathrm{fuse}}\!\left(\mathbf{f}^{(6)}\right)
\right)
\Big].
\end{equation}

The output tokens $\mathbf{U}'$ are obtained by first adding the multimodal positional embeddings $\mathbf{E}^{\mathrm{pos}}_\mathrm{multi}$ to $\mathbf{U}$ and then passing the result through the multimodal Transformer encoder. The refined tokens are then reshaped back ($\mathrm{Reshape}^{-1}$) into a volumetric latent representation, followed by a $1\times1\times1$ convolution, as follows:
\begin{equation}
\mathbf{X}^{\mathrm{inter}}_{6}
=
\mathrm{Conv}^{1\times1\times1}_{\mathrm{dec}}\!\left(
\mathrm{Reshape}^{-1}\!\left(\mathbf{U}'\right)
\right).
\end{equation}

\subsection{Structured Latent Projection Module}
Instead of introducing additional loss terms to explicitly regularize alignment as proposed in \cite{jiang2023understanding}, we propose the Structured latent projection module. This approach injects this inductive bias directly into the architecture by separating the decoder input into a modality-invariant shared component and modality-specific private components, and by routing the private components according to the random modality availability mask. 

Before presenting the details of the structured latent projection module, Theorem 1 is induced from \cite{jiang2023understanding} to explain why enforcing feature alignment can reduce information that is useful for the downstream task and why retaining modality-specific components is beneficial.

\begin{algorithm}
\caption{Multimodal segmentation with structured latent projection under missing modalities}
\label{alg:ours2_short_novel}
\begin{algorithmic}[1]
\Require Multimodal input $\mathbf{X}=[\mathbf{X}_{1},\ldots,\mathbf{X}_{M}]$, availability mask $\mathbf{s}\in\{0,1\}^{B\times M}$
\Ensure Segmentation prediction $\hat{\mathbf{Y}}$

\Statex \textbf{Encoding and cross-modal fusion}
\For{$m=1$ to $M$}
    \State $\big(\mathbf{x}^{(1)}_{m},\ldots,\mathbf{x}^{(6)}_{m}\big)\leftarrow E_m(\mathbf{X}_m)$
    \State $\tilde{\mathbf{x}}^{(l)}_{m}\leftarrow s_{b,m}\odot \mathbf{x}^{(l)}_{m}\quad \forall\, l\in\{1,\ldots,6\}$
\EndFor
\State $\mathbf{f}^{(l)}\leftarrow \Phi^{(l)}\!\big(\tilde{\mathbf{x}}^{(l)}_{1},\ldots,\tilde{\mathbf{x}}^{(l)}_{M}\big)\quad \forall\, l\in\{1,\ldots,6\}$

\Statex \textbf{Intra-modal Transformer}
\For{$m=1$ to $M$}
    \State Compute $\mathbf{Z}_{m}$ as in Eq.~(9)
\EndFor

\Statex \textbf{Inter-modal correlation modeling}
\State $\left[\mathbf{Q}_{m},\mathbf{K}_{m},\mathbf{V}_{m}\right]\leftarrow \mathrm{Conv}^{1\times1\times1}_{qkv,m}\!\big(\mathrm{Reshape}^{-1}(\hat{\mathbf{Z}}_{m})\big)\quad$
\For{$m=1$ to $M$}
    \State $\mathbf{Z}^{\mathrm{corr}}_{m}
\leftarrow
\sum\limits_{j\in\mathcal{M}}
\boldsymbol{\alpha}_{m\leftarrow j}\odot \mathbf{V}_{j}$
    \State $\tilde{\mathbf{Z}}^{\mathrm{corr}}_{m}\leftarrow s_{b,m}\odot \mathbf{Z}^{\mathrm{corr}}_{m}$
    \State $\mathbf{Z}^{\mathrm{fused}}_{m}\leftarrow \mathbf{Z}_{m}+\mathrm{Reshape}\!\big(\tilde{\mathbf{Z}}^{\mathrm{corr}}_{m}\big)$
\EndFor

\Statex \textbf{Multimodal Transformer fusion}
\State $\mathbf{Z}_{\mathrm{multi}}
\leftarrow
\Big[
\mathbf{Z}^{\mathrm{fused}}_{1},
\mathbf{Z}^{\mathrm{fused}}_{2},
\ldots,
\mathbf{Z}^{\mathrm{fused}}_{M}
\Big]$
\State $\mathbf{U}
\leftarrow
\Big[
\mathbf{Z}_{\mathrm{multi}},
\mathrm{Reshape}\!\left(
\mathrm{Conv}^{1\times1\times1}_{\mathrm{fuse}}\!\left(\mathbf{f}^{(6)}\right)
\right)
\Big]$
\State Compute $\mathbf{U}'$ 
\State $\mathbf{X}^{\mathrm{inter}}_{6}\leftarrow \mathrm{Conv}^{1\times1\times1}_{\mathrm{dec}}\!\big(\mathrm{Reshape}^{-1}(\mathbf{U}')\big)$

\Statex \textbf{Structured latent projection and routing}
\State $\mathbf{z}^{\mathrm{sh}}\leftarrow \mathrm{Conv}^{1\times1\times1}_{\mathrm{sh}}\!\big(\mathbf{X}^{\mathrm{inter}}_{6}\big)$
\For{$m=1$ to $M$}
    \State $\mathbf{z}^{\mathrm{pr}}_{m}\leftarrow \mathrm{Conv}^{1\times1\times1}_{\mathrm{pr},m}\!\big(\mathbf{Z}^{\mathrm{corr}}_{m}\big)$
    \State $\tilde{\mathbf{z}}^{\mathrm{pr}}_{m}\leftarrow s_{b,m}\odot \mathbf{z}^{\mathrm{pr}}_{m}$
\EndFor
\State $\mathbf{z}_{6}\leftarrow \big[\mathbf{z}^{\mathrm{sh}},\tilde{\mathbf{z}}^{\mathrm{pr}}_{1},\ldots,\tilde{\mathbf{z}}^{\mathrm{pr}}_{M}\big]$

\Statex \textbf{Decoder}
\State $\hat{\mathbf{Y}}\leftarrow D\!\big(\mathbf{f}^{(1)},\mathbf{f}^{(2)},\mathbf{f}^{(3)},\mathbf{f}^{(4)},\mathbf{z}_{6}\big)$
\State \Return $\hat{\mathbf{Y}}$
\end{algorithmic}
\end{algorithm}

Following we instantiate Theorem 3.1. in \cite{jiang2023understanding}:
\begin{theorem}
Given $M$ modalities with modality encoders $\{g_i(\cdot)\}_{i=1}^{M}$, let $\mathbf{Z}_i=g_i(\mathbf{X}_i)$ denote the encoded feature of modality $i$.
If the multimodal features are perfectly aligned in the feature space, i.e., $\mathbf{Z}_1 = \mathbf{Z}_2 = \cdots = \mathbf{Z}_M$, then
\begin{equation}
\begin{aligned}
\inf_{h}\ \mathbb{E}_{p}\!\left[
\ell_{\mathrm{CE}}\!\big(h(\mathbf{Z}_1,\ldots,\mathbf{Z}_M),\mathbf{Y}\big)
\right]
- \\
\inf_{h'}\ \mathbb{E}_{p}\!\left[
\ell_{\mathrm{CE}}\!\big(h'(\mathbf{X}_1,\ldots,\mathbf{X}_M),\mathbf{Y}\big)
\right]
\ge \Delta_p.
\end{aligned}
\end{equation}
\end{theorem}

Here, $h$ denotes the predictor defined on the encoded features, while $h'$ is the predictor on the original input features.
$\mathbb{E}_p[\cdot]$ denotes expectation under the joint data distribution $p(\mathbf{X}_1,\ldots,\mathbf{X}_M,\mathbf{Y})$ or $p(\mathbf{Z}_1,\ldots,\mathbf{Z}_M,\mathbf{Y})$ and $\ell_{\mathrm{CE}}$ is the cross-entropy loss.
The notation $\inf_{h}$ defines the infimum over all prediction functions $h$.

We define $\Delta_p$ as the difference between the most and the least informative modalities about the target $\mathbf{Y}$:
\begin{equation}
\Delta_p \;=\; \max_{i\in\{1,\ldots,M\}} I(\mathbf{X}_i;\mathbf{Y})\;-\;\min_{i\in\{1,\ldots,M\}} I(\mathbf{X}_i;\mathbf{Y}).
\end{equation}

Assuming any given two random variables $\mathbf{X}$ and $\mathbf{Y}$, $I(\mathbf{X};\mathbf{Y})$ is defined as the Shannon mutual information between $\mathbf{X}$ and $\mathbf{Y}$. Conditional entropy $H(\mathbf{Y}\mid \mathbf{X})$ can be expressed as the optimal expected cross-entropy as follows
\[
H(\mathbf{Y}\mid \mathbf{X}) \;=\; \inf_h \mathbb{E}_p\!\big[\ell_{\mathrm{CE}}(h(\mathbf{X}),\mathbf{Y})\big].
\]
Therefore, for $M$ modalities, Equation (19) can be expressed as
\begin{equation}
\begin{aligned}
H\!\left(\mathbf{Y}\mid \mathbf{Z}_1,\ldots,\mathbf{Z}_M\right)-H\!\left(\mathbf{Y}\mid \mathbf{X}_1,\ldots,\mathbf{X}_M\right)
= \\
I(\mathbf{X}_1,\ldots,\mathbf{X}_M;\mathbf{Y})-I(\mathbf{Z}_1,\ldots,\mathbf{Z}_M;\mathbf{Y})
\;\ge\; \Delta p.
\end{aligned}
\end{equation}

This expression shows that enforcing perfect alignment across multimodal features, e.g., $\mathbf{Z}_1=\mathbf{Z}_2=\cdots=\mathbf{Z}_M$, can increase the optimal cross-entropy loss $\ell_{\mathrm{CE}}$ by at least $\Delta_p$, since the aligned representation cannot retain modality-specific information~\cite{farnia2016minimax, zhao2022fundamental}.

This provides a theoretical motivation that using only a single shared latent may be suboptimal, especially when different modalities contribute unequal information for the downstream task. To avoid this issue, the fused representation is decomposed into a modality-invariant shared component $\mathbf{z}^{\mathrm{sh}}$ and modality-specific private components $\{\mathbf{z}^{\mathrm{pr}}_m\}_{m\in\mathcal{M}}$. During decoding, $\mathbf{z}^{\mathrm{sh}}$ is always provided, while private components are routed according to the random modality mask, namely $\tilde{\mathbf{z}}^{\mathrm{pr}}_m = s_{b,m}\odot \mathbf{z}^{\mathrm{pr}}_m$. This approach preserves discriminative modality-specific cues in the full modality case.

\begin{figure*}[t]
\centering
\includegraphics[width=\textwidth]{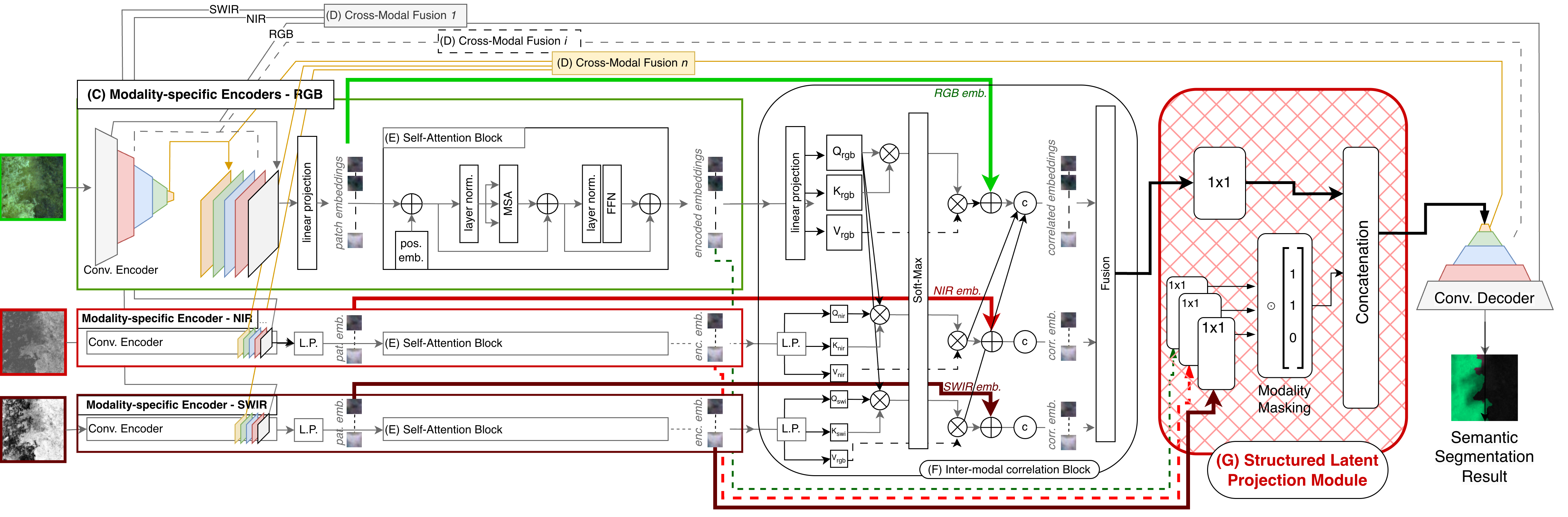}
\caption{Framework of the proposed CBC-SLP model that learns structured multimodal feature representations for robust semantic segmentation under missing or full modalities. CBC-SLP is based on a structured latent projection design that explicitly models shared and modality-specific representations before decoding.}
\label{fig:latent}
\end{figure*}

Let $\mathbf{X}^{\mathrm{inter}}_{6}$ denote the inter-modal fused latent produced by the inter-modal correlation modeling and the subsequent multimodal Transformer. For each modality $m\in\mathcal{M}$, let $\mathbf{Z}^{\mathrm{corr}}_{m}$ denote the ungated modality-specific representation obtained after inter-modal correlation aggregation. 

As illustrated in Fig.~\ref{fig:latent}, the structured latent projection module constructs a modality-invariant shared latent space $\mathbf{z}^{\mathrm{sh}}$ from $\mathbf{X}^{\mathrm{inter}}_{6}$ and a set of modality-specific private latent spaces $\{\mathbf{z}^{\mathrm{pr}}_{m}\}_{m\in\mathcal{M}}$ from $\{\mathbf{Z}^{\mathrm{corr}}_{m}\}_{m\in\mathcal{M}}$, and forms the decoder input by routing the private components through multiplicative gating using the modality availability mask.

\paragraph{Shared Latent Projection}
The shared component is obtained by projecting the inter-modal fused latent $\mathbf{X}^{\mathrm{inter}}_{6}$ with a $1\times1\times1$ convolution as follows:

\begin{equation}
\mathbf{z}^{\mathrm{sh}}
=
\mathrm{Conv}^{1\times1\times1}_{\mathrm{sh}}\!\left(\mathbf{X}^{\mathrm{inter}}_{6}\right).
\end{equation}
where $\mathbf{z}^{\mathrm{sh}}\in\mathbb{R}^{B\times C_z\times D_t\times H_t\times W_t}$ is the resulting shared latent with $C_z$ channels.

\paragraph{Private Latent Projections}
For each modality $m\in\mathcal{M}$, the private component is obtained by projecting the modality-specific correlation representation $\mathbf{Z}^{\mathrm{corr}}_{m}$ with a $1\times1\times1$ convolution as follows:
\begin{equation}
\mathbf{z}^{\mathrm{pr}}_{m}
=
\mathrm{Conv}^{1\times1\times1}_{\mathrm{pr},m}\!\left(\mathbf{Z}^{\mathrm{corr}}_{m}\right).
\end{equation}
where $\mathbf{z}^{\mathrm{pr}}_{m}\in\mathbb{R}^{B\times C_z\times D_t\times H_t\times W_t}$ is the resulting private latent with the same channel width $C_z$ as the shared latent space.

\paragraph{Latent Space Routing Using the Modality Mask}
The private components are routed to the decoder by multiplicative gating as follows:

\begin{equation}
\tilde{\mathbf{z}}^{\mathrm{pr}}_{m}
=
s_{b,m}\odot \mathbf{z}^{\mathrm{pr}}_{m}.
\end{equation}
where $s_{b,m}$ is broadcast over the channel and spatial dimensions of $\mathbf{z}^{\mathrm{pr}}_{m}$. This routing yields $\tilde{\mathbf{z}}^{\mathrm{pr}}_{m}=\mathbf{0}$ when modality $m$ is missing and preserves modality-specific cues when modality $m$ is available.

\paragraph{Decoder Input Composition}
The latent tensor $\mathbf{z}_{6}\in\mathbb{R}^{B\times((M+1)C_{z})\times D_t\times H_t\times W_t}$ provided to the decoder is formed by concatenating the shared component with all routed private components along the channel dimension when all modalities are available as follows:
\begin{equation}
\mathbf{z}_{6}
=
\Big[
\mathbf{z}^{\mathrm{sh}},
\tilde{\mathbf{z}}^{\mathrm{pr}}_{1},
\tilde{\mathbf{z}}^{\mathrm{pr}}_{2},
\ldots,
\tilde{\mathbf{z}}^{\mathrm{pr}}_{M}
\Big].
\end{equation}
where $\big[\cdot\big]$ denotes concatenation along the channel dimension.

\paragraph{Full and missing modality behavior}
For a sample index $b\in\{1,\ldots,B\}$, the full modality case satisfies $\sum_{m=1}^{M} s_{b,m}=M$, and the decoder input reduces to the following:

\begin{equation}
\mathbf{z}^{(b)}_{6}
=
\Big[
\mathbf{z}^{\mathrm{sh}(b)},
\mathbf{z}^{\mathrm{pr}(b)}_{1},
\mathbf{z}^{\mathrm{pr}(b)}_{2},
\ldots,
\mathbf{z}^{\mathrm{pr}(b)}_{M}
\Big].
\end{equation}

For the missing modality case, $\sum_{m=1}^{M} s_{b,m}<M$, and the decoder receives the following:

\begin{equation}
\mathbf{z}^{(b)}_{6}
=
\Big[
\mathbf{z}^{\mathrm{sh}(b)},
\tilde{\mathbf{z}}^{\mathrm{pr}(b)}_{1},
\tilde{\mathbf{z}}^{\mathrm{pr}(b)}_{2},
\ldots,
\tilde{\mathbf{z}}^{\mathrm{pr}(b)}_{M}
\Big],
\end{equation}
where $\tilde{\mathbf{z}}^{\mathrm{pr}(b)}_{m}=\mathbf{0}$ holds for every modality index $m$ such that $s_{b,m}=0$. This formulation retains modality-specific cues when all modalities are available and prevents private signals from missing modalities from being injected into the decoder.

\subsection{Decoder}
We fuse the first 4 skip features originating from the modality-specific encoders and the structured latent tensor $\mathbf{z}_{6}$ constructed in the previous section. Our lightweight decoder produces the final prediction as follows:
\begin{equation}
\hat{\mathbf{Y}}
=
D\!\left(
\mathbf{f}^{(1)},
\mathbf{f}^{(2)},
\mathbf{f}^{(3)},
\mathbf{f}^{(4)},
\mathbf{z}_{6}
\right),
\end{equation}
where $D(\cdot)$ denotes a convolutional decoder that upsamples $\mathbf{z}_{6}$ while integrating the skip features, and outputs $\hat{\mathbf{Y}}$ by applying a sigmoid function to the final logits. The complete pipeline is summarized in Algorithm~\ref{alg:ours2_short_novel}.

\section{Experiments}
In this section, we conduct thorough comparative experiments to demonstrate the effectiveness of the proposed CBC-SLP model under both missing and full modality settings.

\subsection{Implementation Details}
All experiments are implemented in PyTorch and trained on a workstation equipped with an NVIDIA RTX 6000 Ada GPU. During training, the Adam optimizer is used with an initial learning rate of $1\times10^{-4}$, a mini-batch size of 8, and a total of 70 epochs. A learning rate decay schedule is applied, reducing the learning rate by 9\% every 5 epochs. A 5-fold cross-validation protocol is adopted for robust evaluation. For both training and validation, the loss function is  binary cross-entropy (BCE). For fair comparison, all state-of-the-art models are trained from scratch using the same hyperparameter settings and the same image set split protocol.

\begin{figure*}[!h]
    \centering
    \includegraphics[width=\textwidth]{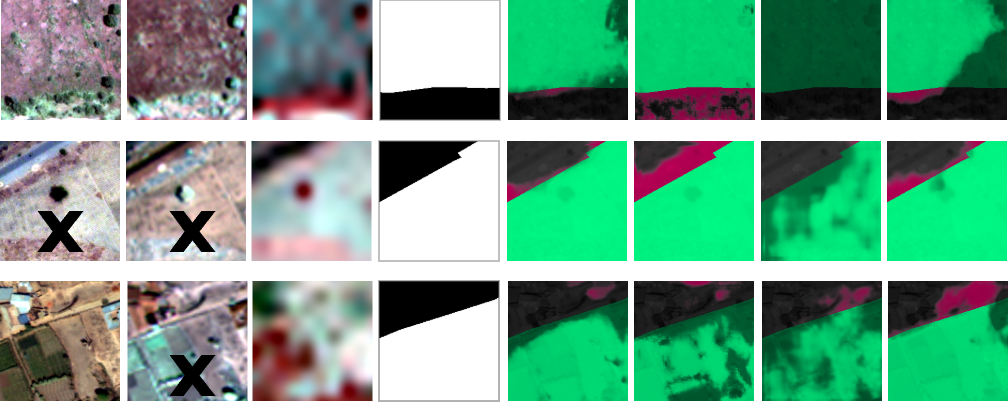}
    \caption{Qualitative semantic segmentation comparison on the DSTL image set under different modality-availability settings. From left to right: (a) RGB image, (b) NIR image, (c) SWIR image, (d) ground truth, (e) CBC-SLP (ours), (f) CBC (baseline), (g) CMX, (h) M3L. Row 1 uses all modalities; Row 2 uses only SWIR; Row 3 excludes NIR.}
    \label{fig:dstl_visual}
\end{figure*}

\begin{figure*}[!h]
    \centering
    \includegraphics[width=\textwidth]{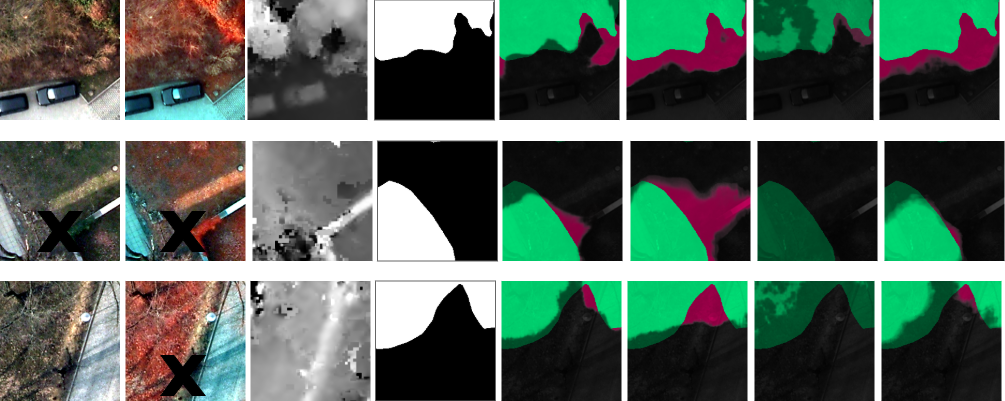}
    \caption{Qualitative semantic segmentation comparison on the Potsdam image set under different modality-availability settings. From left to right: (a) RGB image, (b) IRRG image, (c) DSM image, (d) ground truth, (e) CBC-SLP (ours), (f) CBC (baseline), (g) CMX, (h) M3L. Row 1 uses all modalities; Row 2 uses only DSM; Row 3 excludes IRRG.}
    \label{fig:potsdam_visual}
\end{figure*}

\begin{figure*}[!h]
    \centering
    \includegraphics[width=\textwidth]{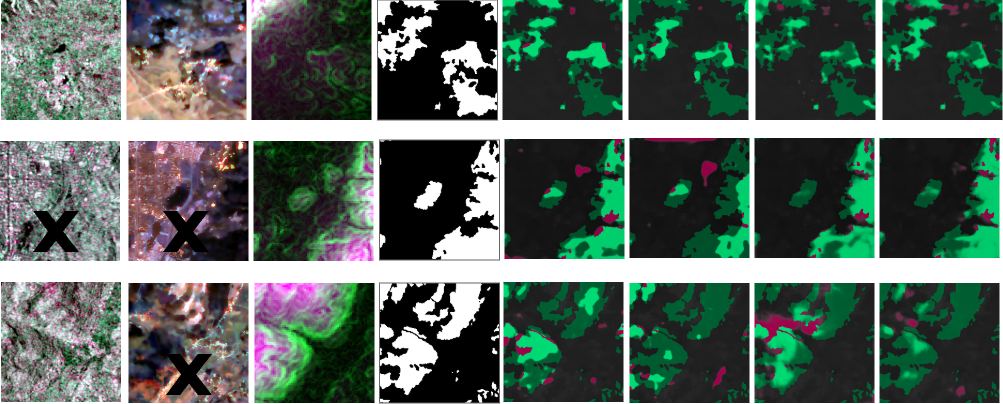}
    \caption{Qualitative semantic segmentation comparison on the Hunan image set under different modality-availability settings. From left to right: (a) SAR image, (b) MSI image, (c) DEM image, (d) ground truth, (e) CBC-SLP (ours), (f) CBC (baseline), (g) CMX, (h) M3L. Row 1 uses all modalities; Row 2 excludes MSI; Row 3 uses only DEM.}
    \label{fig:hunan_visual}
\end{figure*}

\subsection{Evaluation Metrics}
We evaluate performance using the Intersection over Union (IoU) and $F_1$ score metrics \cite{cho2024weighted} under all modality availability scenarios. Since we perform binary semantic segmentation for a selected target class in each image set, the metrics are computed for that class only. 
A higher IoU indicates a larger pixel-wise overlap between the predicted and ground-truth target regions relative to their union. For $F_1$, a higher value indicates a better balance between misses and false alarms while rewarding correct detections.

\subsection{Comparison With State-of-The-Art Multimodal Models}

In Table~\ref{tab:dstl_sota_comparison}, we compare the performance of our proposed model with that of other state-of-the-art multimodal models such as MultiSenseSeg~\cite{wang2024multisenseseg}, CFFormer~\cite{zhao2024cfformer}, CMX~\cite{zhang2023cmx}, CMNeXt~\cite{zhang2023delivering}, Dformerv2-S~\cite{yin2025dformerv2}, M3L~\cite{maheshwari2024missing}, MMANet~\cite{wei2023mmanet} on the DSTL test set. Compared to others, our CBC-SLP model achieves better state-of-the-art performance for all modality settings. Specifically, when all modalities are available, our model outperforms the previously best-performing M3L and MMANet models by approximately 1.6\% and 2.7\% IoU, respectively.

Achieving higher performance in cases where two modalities are missing remains another goal, in addition to the case when all modalities are available. When both RGB and NIR modalities are missing, our model achieves an IoU of 0.865, which is 1.2\% and 1.3\% higher than Dformerv2-S and CMNeXt, respectively. The overall improvements show that by incorporating modality-specific private components during training with a random modality availability mask, our CBC-SLP model gains a better capacity to handle both missing and full modality cases.

Similarly, as shown in Table~\ref{tab:potsdam_sota_comparison}, test set results on Potsdam for the tree class further confirm the effectiveness of the proposed model. Compared to M3L, when all modalities are available, our model improves IoU by 4.4\%. In particular, for the challenging only DSM available case, our model achieves an IoU of 0.576, surpassing the M3L model by over 0.3\% IoU. Considering the case when only the IRRG modality is available, the model improves the IoU of Dformerv2-S by 4.3\%. Therefore, CBC-SLP model maintains a robust performance in all scenarios.

As presented in Table~\ref{tab:hunan_sota_comparison}, on the Hunan test set, our model outperforms for almost all cases. Compared to CMX, it achieves better segmentation performance when all modalities are available, with an IoU gain of 0.8\%, due to its novel shared and modality-specific latent decomposition. When compared to MMANet, CBC-SLP improves IoU by 1.3\% when only the MSI modality is used. However, CBC-SLP performs slightly worse than others in only one case, namely when only the DEM modality is available. Because DEM features are less discriminative for forest segmentation than shared features, in this case the model may be relying more on modality-invariant shared representations than on private DEM features.

\begin{table*}[t]
\centering
\scriptsize
\setlength{\tabcolsep}{2.8pt}
\renewcommand{\arraystretch}{1.12}

\caption{COMPARISON OF {THE} PROPOSED MODEL WITH STATE-OF-THE-ART APPROACHES ON DSTL TEST SET}
\label{tab:dstl_sota_comparison}

\resizebox{\textwidth}{!}{%
\begin{tabular}{c|l|cc|cc|cc|cc|cc|cc|cc}
\Xhline{1.1pt}
\textbf{Image Set} &
\multicolumn{1}{c|}{\textbf{Model}} &
\multicolumn{2}{c|}{\textbf{Full}} &
\multicolumn{2}{c|}{\textbf{RGB missing}} &
\multicolumn{2}{c|}{\textbf{NIR missing}} &
\multicolumn{2}{c|}{\textbf{SWIR missing}} &
\multicolumn{2}{c|}{\textbf{RGB only}} &
\multicolumn{2}{c|}{\textbf{NIR only}} &
\multicolumn{2}{c}{\textbf{SWIR only}} \\
\cline{3-16}
& &
\textbf{IoU} & \textbf{F$_1$} &
\textbf{IoU} & \textbf{F$_1$} &
\textbf{IoU} & \textbf{F$_1$} &
\textbf{IoU} & \textbf{F$_1$} &
\textbf{IoU} & \textbf{F$_1$} &
\textbf{IoU} & \textbf{F$_1$} &
\textbf{IoU} & \textbf{F$_1$} \\
\Xhline{0.9pt}

\multirow{8}{*}{\cellcolor[gray]{0.9}\rotatebox{90}{\textbf{DSTL}}}
& MultiSenseSeg~\cite{wang2024multisenseseg}
& 0.863$\pm$0.312 & 0.875$\pm$0.304
& 0.855$\pm$0.321 & 0.868$\pm$0.313
& 0.860$\pm$0.317 & 0.872$\pm$0.310
& 0.858$\pm$0.318 & 0.871$\pm$0.310
& 0.849$\pm$0.327 & 0.862$\pm$0.320
& 0.836$\pm$0.341 & 0.849$\pm$0.335
& 0.850$\pm$0.328 & 0.862$\pm$0.320 \\

& CFFormer~\cite{zhao2024cfformer}
& 0.849$\pm$0.336 & 0.858$\pm$0.331
& 0.850$\pm$0.333 & 0.860$\pm$0.327
& 0.847$\pm$0.340 & 0.856$\pm$0.335
& 0.824$\pm$0.360 & 0.834$\pm$0.355
& 0.816$\pm$0.371 & 0.823$\pm$0.368
& 0.820$\pm$0.363 & 0.830$\pm$0.357
& 0.844$\pm$0.339 & 0.854$\pm$0.333 \\

& CMX~\cite{zhang2023cmx}
& 0.860$\pm$0.324 & 0.869$\pm$0.319
& 0.853$\pm$0.332 & 0.862$\pm$0.326
& 0.859$\pm$0.325 & 0.868$\pm$0.320
& 0.853$\pm$0.333 & 0.862$\pm$0.328
& 0.854$\pm$0.331 & 0.863$\pm$0.325
& 0.806$\pm$0.377 & 0.815$\pm$0.371
& 0.849$\pm$0.335 & 0.859$\pm$0.328 \\

& CMNeXt~\cite{zhang2023delivering}
& 0.862$\pm$0.320 & 0.872$\pm$0.314
& 0.862$\pm$0.319 & 0.872$\pm$0.313
& 0.855$\pm$0.330 & 0.864$\pm$0.326
& 0.862$\pm$0.320 & 0.872$\pm$0.314
& 0.843$\pm$0.342 & 0.852$\pm$0.336
& 0.853$\pm$0.327 & 0.864$\pm$0.320
& 0.852$\pm$0.323 & 0.866$\pm$0.314 \\

& Dformerv2-S~\cite{yin2025dformerv2}
& 0.856$\pm$0.328 & 0.865$\pm$0.322
& 0.858$\pm$0.325 & 0.868$\pm$0.319
& 0.855$\pm$0.328 & 0.865$\pm$0.322
& 0.855$\pm$0.333 & 0.862$\pm$0.330
& 0.854$\pm$0.332 & 0.862$\pm$0.328
& 0.857$\pm$0.326 & 0.867$\pm$0.321
& 0.853$\pm$0.328 & 0.864$\pm$0.320 \\

& M3L~\cite{maheshwari2024missing}
& 0.887$\pm$0.273 & 0.902$\pm$0.260
& 0.878$\pm$0.282 & 0.895$\pm$0.268
& 0.871$\pm$0.298 & 0.885$\pm$0.285
& 0.888$\pm$0.272 & 0.903$\pm$0.261
& 0.874$\pm$0.302 & 0.885$\pm$0.301
& 0.876$\pm$0.283 & 0.893$\pm$0.269
& 0.849$\pm$0.321 & 0.865$\pm$0.310 \\

& MMANet~\cite{wei2023mmanet}
& 0.876$\pm$0.295 & 0.889$\pm$0.284
& 0.875$\pm$0.296 & 0.888$\pm$0.285
& 0.865$\pm$0.315 & 0.875$\pm$0.310
& 0.879$\pm$0.291 & 0.892$\pm$0.280
& 0.865$\pm$0.315 & 0.875$\pm$0.310
& 0.873$\pm$0.295 & 0.887$\pm$0.283
& 0.851$\pm$0.329 & 0.862$\pm$0.322 \\
\cline{2-16}
& \textbf{CBC-SLP (ours)}
& \textbf{0.903$\pm$0.255} & \textbf{0.916$\pm$0.242}
& \textbf{0.901$\pm$0.256} & \textbf{0.915$\pm$0.241}
& \textbf{0.893$\pm$0.269} & \textbf{0.907$\pm$0.255}
& \textbf{0.902$\pm$0.256} & \textbf{0.916$\pm$0.241}
& \textbf{0.876$\pm$0.300} & \textbf{0.887$\pm$0.293}
& \textbf{0.896$\pm$0.265} & \textbf{0.910$\pm$0.251}
& \textbf{0.865$\pm$0.306} & \textbf{0.879$\pm$0.294} \\
\Xhline{1.1pt}
\end{tabular}%
}

\vspace{0.6mm}
\footnotesize
Random modality dropout training setting is used. The best scores under each modality availability combination are marked as bold.
\end{table*}


\begin{table*}[t]
\centering
\scriptsize
\setlength{\tabcolsep}{2.8pt}
\renewcommand{\arraystretch}{1.12}

\caption{COMPARISON OF {THE} PROPOSED MODEL WITH STATE-OF-THE-ART APPROACHES ON POTSDAM TEST SET}
\label{tab:potsdam_sota_comparison}

\resizebox{\textwidth}{!}{%
\begin{tabular}{c|l|cc|cc|cc|cc|cc|cc|cc}
\Xhline{1.1pt}
\textbf{Image Set} &
\multicolumn{1}{c|}{\textbf{Model}} &
\multicolumn{2}{c|}{\textbf{Full}} &
\multicolumn{2}{c|}{\textbf{RGB missing}} &
\multicolumn{2}{c|}{\textbf{IRRG missing}} &
\multicolumn{2}{c|}{\textbf{DSM missing}} &
\multicolumn{2}{c|}{\textbf{RGB only}} &
\multicolumn{2}{c|}{\textbf{IRRG only}} &
\multicolumn{2}{c}{\textbf{DSM only}} \\
\cline{3-16}
& &
\textbf{IoU} & \textbf{F$_1$} &
\textbf{IoU} & \textbf{F$_1$} &
\textbf{IoU} & \textbf{F$_1$} &
\textbf{IoU} & \textbf{F$_1$} &
\textbf{IoU} & \textbf{F$_1$} &
\textbf{IoU} & \textbf{F$_1$} &
\textbf{IoU} & \textbf{F$_1$} \\
\Xhline{0.9pt}

\multirow{8}{*}{\cellcolor[gray]{0.9}\rotatebox{90}{\textbf{Potsdam}}}
& MultiSenseSeg~\cite{wang2024multisenseseg}
& 0.644$\pm$0.399 & 0.690$\pm$0.389
& 0.637$\pm$0.403 & 0.683$\pm$0.393
& 0.639$\pm$0.401 & 0.685$\pm$0.392
& 0.628$\pm$0.403 & 0.675$\pm$0.391
& 0.593$\pm$0.421 & 0.636$\pm$0.413
& 0.608$\pm$0.413 & 0.653$\pm$0.403
& 0.548$\pm$0.442 & 0.584$\pm$0.439 \\

& CFFormer~\cite{zhao2024cfformer}
& 0.546$\pm$0.428 & 0.594$\pm$0.410
& 0.538$\pm$0.434 & 0.582$\pm$0.420
& 0.506$\pm$0.456 & 0.538$\pm$0.446
& 0.543$\pm$0.429 & 0.591$\pm$0.411
& 0.492$\pm$0.464 & 0.518$\pm$0.454
& 0.532$\pm$0.438 & 0.574$\pm$0.424
& 0.449$\pm$0.491 & 0.453$\pm$0.491 \\

& CMX~\cite{zhang2023cmx}
& 0.541$\pm$0.426 & 0.590$\pm$0.408
& 0.532$\pm$0.434 & 0.577$\pm$0.418
& 0.516$\pm$0.448 & 0.553$\pm$0.436
& 0.539$\pm$0.426 & 0.589$\pm$0.408
& 0.519$\pm$0.445 & 0.558$\pm$0.430
& 0.526$\pm$0.435 & 0.571$\pm$0.421
& 0.444$\pm$0.492 & 0.448$\pm$0.493 \\

& CMNeXt~\cite{zhang2023delivering}
& 0.686$\pm$0.370 & 0.737$\pm$0.354
& 0.676$\pm$0.377 & 0.726$\pm$0.361
& 0.628$\pm$0.405 & 0.675$\pm$0.394
& 0.665$\pm$0.383 & 0.715$\pm$0.368
& 0.548$\pm$0.444 & 0.583$\pm$0.437
& 0.650$\pm$0.391 & 0.700$\pm$0.378
& 0.568$\pm$0.433 & 0.607$\pm$0.428 \\

& Dformerv2-S~\cite{yin2025dformerv2}
& 0.716$\pm$0.364 & 0.760$\pm$0.355
& 0.712$\pm$0.365 & 0.757$\pm$0.355
& 0.682$\pm$0.387 & 0.724$\pm$0.381
& 0.712$\pm$0.366 & 0.757$\pm$0.356
& 0.672$\pm$0.393 & 0.714$\pm$0.386
& 0.707$\pm$0.368 & 0.752$\pm$0.358
& 0.483$\pm$0.479 & 0.498$\pm$0.478 \\

& M3L~\cite{maheshwari2024missing}
& 0.722$\pm$0.345 & 0.773$\pm$0.330
& 0.705$\pm$0.358 & 0.756$\pm$0.343
& 0.699$\pm$0.361 & 0.750$\pm$0.346
& 0.708$\pm$0.356 & 0.759$\pm$0.342
& 0.679$\pm$0.372 & 0.730$\pm$0.359
& 0.687$\pm$0.370 & 0.737$\pm$0.357
& 0.573$\pm$0.433 & 0.611$\pm$0.429 \\

& MMANet~\cite{wei2023mmanet}
& 0.699$\pm$0.366 & 0.748$\pm$0.353
& 0.692$\pm$0.370 & 0.741$\pm$0.357
& 0.672$\pm$0.381 & 0.720$\pm$0.370
& 0.674$\pm$0.382 & 0.722$\pm$0.370
& 0.622$\pm$0.409 & 0.667$\pm$0.400
& 0.673$\pm$0.379 & 0.722$\pm$0.366
& 0.561$\pm$0.443 & 0.595$\pm$0.440 \\
\cline{2-16}

& \textbf{CBC-SLP (ours)}
& \textbf{0.766$\pm$0.325} & \textbf{0.811$\pm$0.312}
& \textbf{0.759$\pm$0.331} & \textbf{0.803$\pm$0.317}
& \textbf{0.713$\pm$0.362} & \textbf{0.759$\pm$0.351}
& \textbf{0.761$\pm$0.329} & \textbf{0.806$\pm$0.316}
& \textbf{0.691$\pm$0.376} & \textbf{0.737$\pm$0.366}
& \textbf{0.750$\pm$0.337} & \textbf{0.795$\pm$0.324}
& \textbf{0.576$\pm$0.432} & \textbf{0.614$\pm$0.427} \\
\Xhline{1.1pt}
\end{tabular}%
}

\vspace{0.6mm}
\footnotesize
Random modality dropout training setting is used. The best scores under each modality availability combination are marked as bold.
\end{table*}


\begin{table*}[t]
\centering
\scriptsize
\setlength{\tabcolsep}{2.8pt}
\renewcommand{\arraystretch}{1.12}

\caption{COMPARISON OF OUR PROPOSED MODEL WITH STATE-OF-THE-ART APPROACHES ON HUNAN TEST SET}
\label{tab:hunan_sota_comparison}

\resizebox{\textwidth}{!}{%
\begin{tabular}{c|l|cc|cc|cc|cc|cc|cc|cc}
\Xhline{1.1pt}
\textbf{Image Set} &
\multicolumn{1}{c|}{\textbf{Model}} &
\multicolumn{2}{c|}{\textbf{Full}} &
\multicolumn{2}{c|}{\textbf{SAR missing}} &
\multicolumn{2}{c|}{\textbf{MSI missing}} &
\multicolumn{2}{c|}{\textbf{DEM missing}} &
\multicolumn{2}{c|}{\textbf{SAR only}} &
\multicolumn{2}{c|}{\textbf{MSI only}} &
\multicolumn{2}{c}{\textbf{DEM only}} \\
\cline{3-16}
& &
\textbf{IoU} & \textbf{F$_1$} &
\textbf{IoU} & \textbf{F$_1$} &
\textbf{IoU} & \textbf{F$_1$} &
\textbf{IoU} & \textbf{F$_1$} &
\textbf{IoU} & \textbf{F$_1$} &
\textbf{IoU} & \textbf{F$_1$} &
\textbf{IoU} & \textbf{F$_1$} \\
\Xhline{0.9pt}

\multirow{8}{*}{\cellcolor[gray]{0.9}\rotatebox{90}{\textbf{Hunan}}}
& MultiSenseSeg~\cite{wang2024multisenseseg}
& 0.614$\pm$0.320 & 0.698$\pm$0.322
& 0.611$\pm$0.335 & 0.689$\pm$0.336
& 0.554$\pm$0.334 & 0.640$\pm$0.347
& 0.599$\pm$0.309 & 0.692$\pm$0.303
& 0.509$\pm$0.288 & 0.618$\pm$0.303
& 0.573$\pm$0.334 & 0.658$\pm$0.336
& 0.519$\pm$0.369 & 0.592$\pm$0.382 \\

& CFFormer~\cite{zhao2024cfformer}
& 0.622$\pm$0.337 & 0.698$\pm$0.335
& 0.612$\pm$0.342 & 0.688$\pm$0.342
& 0.582$\pm$0.323 & 0.670$\pm$0.326
& 0.611$\pm$0.342 & 0.687$\pm$0.338
& 0.500$\pm$0.296 & 0.607$\pm$0.309
& 0.607$\pm$0.346 & 0.682$\pm$0.344
& 0.534$\pm$0.349 & 0.616$\pm$0.360 \\

& CMX~\cite{zhang2023cmx}
& 0.651$\pm$0.306 & 0.733$\pm$0.300
& 0.625$\pm$0.325 & 0.705$\pm$0.321
& 0.555$\pm$0.328 & 0.644$\pm$0.339
& 0.639$\pm$0.307 & 0.723$\pm$0.304
& 0.476$\pm$0.313 & 0.580$\pm$0.315
& 0.599$\pm$0.326 & 0.683$\pm$0.325
& 0.509$\pm$0.340 & 0.598$\pm$0.343 \\

& CMNeXt~\cite{zhang2023delivering}
& 0.608$\pm$0.341 & 0.687$\pm$0.333
& 0.580$\pm$0.360 & 0.654$\pm$0.355
& 0.523$\pm$0.337 & 0.612$\pm$0.342
& 0.605$\pm$0.341 & 0.684$\pm$0.333
& 0.518$\pm$0.315 & 0.618$\pm$0.316
& 0.576$\pm$0.361 & 0.650$\pm$0.359
& 0.442$\pm$0.369 & 0.519$\pm$0.374 \\

& Dformerv2-S~\cite{yin2025dformerv2}
& 0.612$\pm$0.328 & 0.694$\pm$0.324
& 0.600$\pm$0.331 & 0.683$\pm$0.328
& 0.529$\pm$0.358 & 0.607$\pm$0.367
& 0.597$\pm$0.338 & 0.677$\pm$0.339
& 0.527$\pm$0.353 & 0.607$\pm$0.364
& 0.587$\pm$0.341 & 0.667$\pm$0.342
& 0.504$\pm$0.361 & 0.582$\pm$0.372 \\

& M3L~\cite{maheshwari2024missing}
& 0.603$\pm$0.323 & 0.687$\pm$0.332
& 0.618$\pm$0.331 & 0.697$\pm$0.333
& 0.542$\pm$0.334 & 0.629$\pm$0.346
& 0.581$\pm$0.322 & 0.669$\pm$0.328
& 0.489$\pm$0.328 & 0.585$\pm$0.336
& 0.596$\pm$0.319 & 0.684$\pm$0.319
& \textbf{0.556$\pm$0.349} & \textbf{0.634$\pm$0.363} \\

& MMANet~\cite{wei2023mmanet}
& 0.650$\pm$0.307 & 0.732$\pm$0.306
& 0.617$\pm$0.321 & 0.701$\pm$0.316
& 0.572$\pm$0.321 & 0.661$\pm$0.329
& 0.647$\pm$0.302 & 0.731$\pm$0.299
& 0.557$\pm$0.329 & 0.644$\pm$0.341
& 0.603$\pm$0.315 & 0.693$\pm$0.306
& 0.526$\pm$0.350 & 0.609$\pm$0.358 \\
\cline{2-16}

& \textbf{CBC-SLP (ours)}
& \textbf{0.659$\pm$0.299} & \textbf{0.743$\pm$0.287}
& \textbf{0.640$\pm$0.304} & \textbf{0.727$\pm$0.295}
& \textbf{0.585$\pm$0.343} & \textbf{0.660$\pm$0.362}
& \textbf{0.652$\pm$0.300} & \textbf{0.737$\pm$0.291}
& \textbf{0.569$\pm$0.298} & \textbf{0.668$\pm$0.302}
& \textbf{0.616$\pm$0.313} & \textbf{0.704$\pm$0.306}
& 0.510$\pm$0.334 & 0.600$\pm$0.345 \\
\Xhline{1.1pt}
\end{tabular}%
}

\vspace{0.6mm}
\footnotesize
Random modality dropout training setting is used. The best scores under each modality availability combination are marked as bold.
\end{table*}

\begin{figure}[h]
\centering

\subfloat[]{%
  \includegraphics[width=0.4\textwidth]{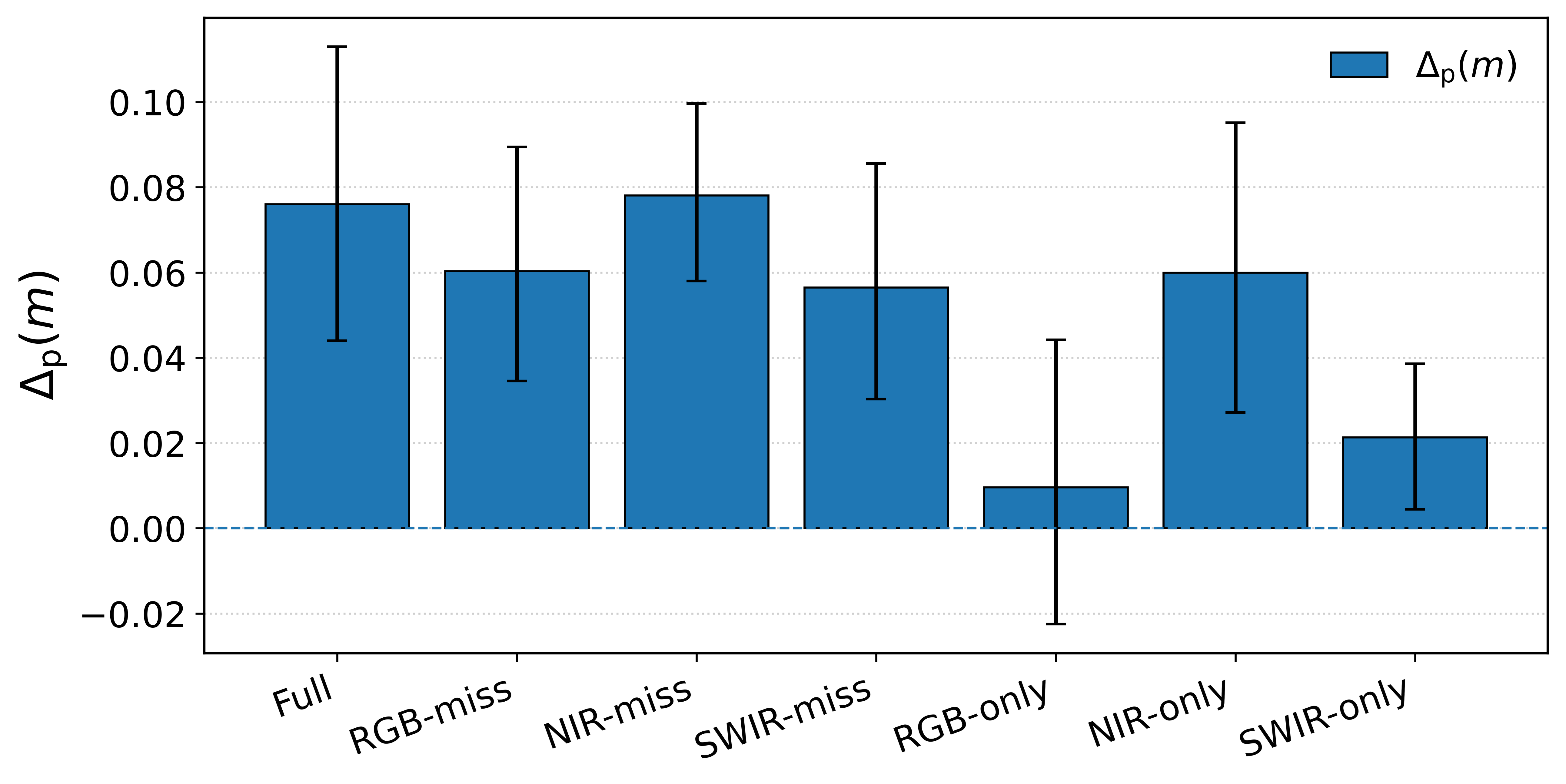}%
  \label{fig:gap_dstl}%
}\par\vspace{0.5mm}

\subfloat[]{%
  \includegraphics[width=0.4\textwidth]{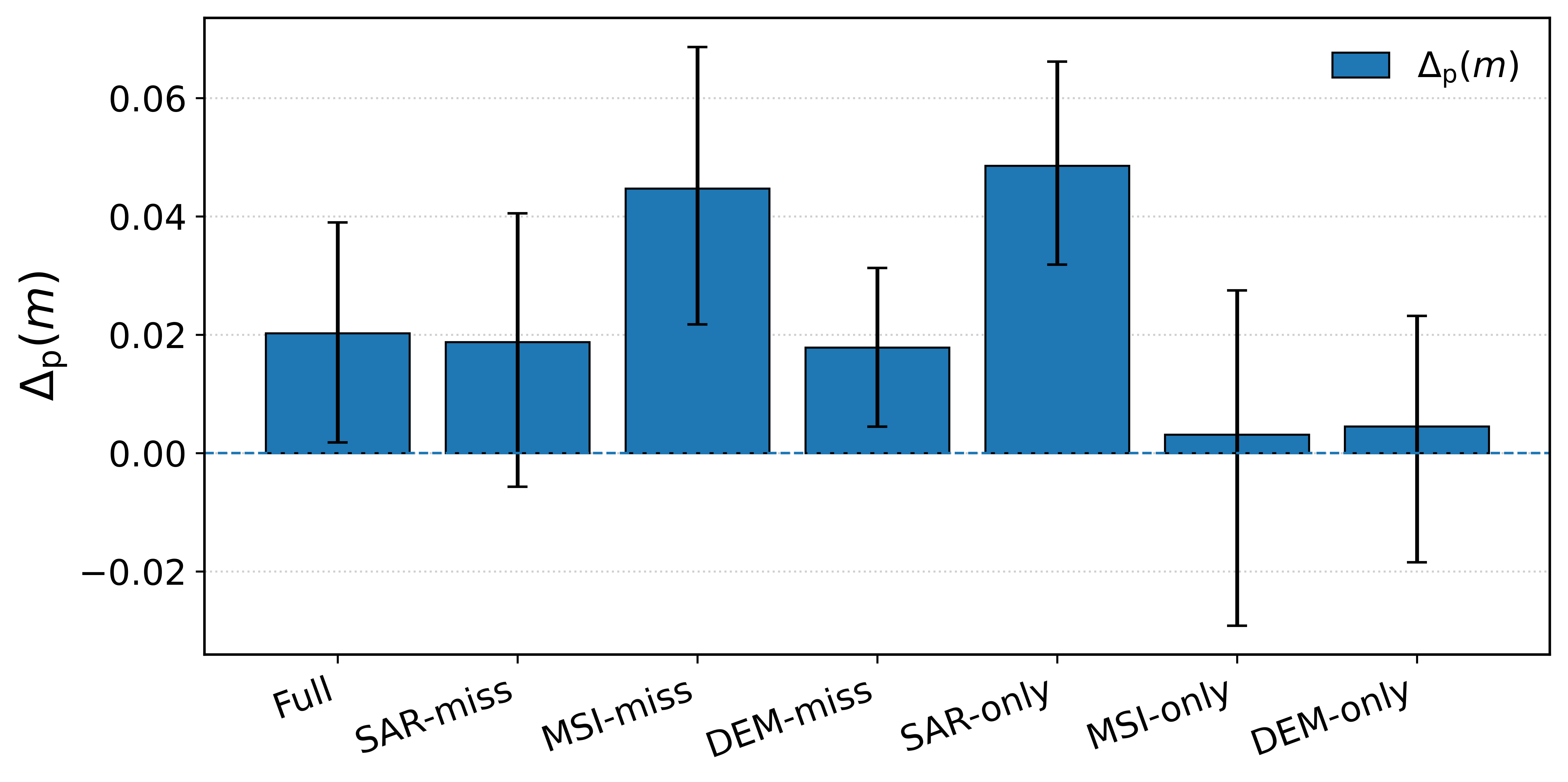}%
  \label{fig:gap_hunan}%
}\par\vspace{0.5mm}

\subfloat[]{%
  \includegraphics[width=0.4\textwidth]{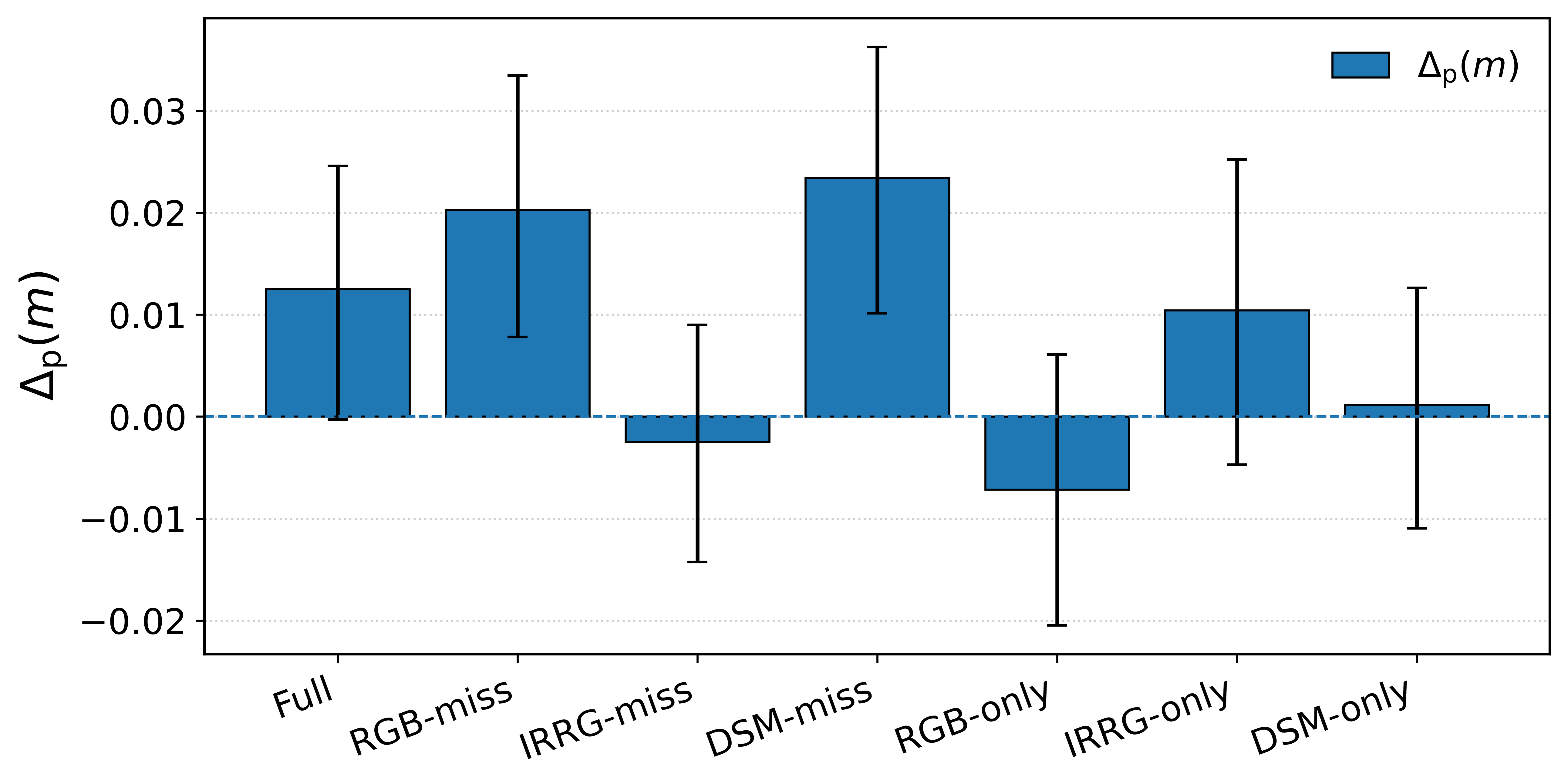}%
  \label{fig:gap_potsdam}%
}

\caption{Information gap across different modality availability scenarios:
(a) DSTL, (b) Hunan, (c) Potsdam.}
\label{fig:gap}
\end{figure}


\begin{table*}[t]
\centering
\caption{COMPARISON OF OUR PROPOSED MODEL WITH BASELINE ON DSTL TEST SET FOR CROP CLASS}
\label{tab:dstl_crop_perf}
\setlength{\tabcolsep}{4.8pt}
\renewcommand{\arraystretch}{1.15}
\begin{tabular}{ccc cccc cccc}
\toprule
\multicolumn{3}{c}{\textbf{Modalities}} &
\multicolumn{4}{c}{\textbf{CBC (baseline)}} &
\multicolumn{4}{c}{\textbf{CBC-SLP (ours)}} \\
\cmidrule(lr){1-3}\cmidrule(lr){4-7}\cmidrule(lr){8-11}
\textbf{RGB} & \textbf{NIR} & \textbf{SWIR} &
\multicolumn{2}{c}{\makecell{\textbf{No modality}\\\textbf{dropout}}} &
\multicolumn{2}{c}{\makecell{\textbf{Random modality}\\\textbf{dropout}}} &
\multicolumn{2}{c}{\makecell{\textbf{No modality}\\\textbf{dropout}}} &
\multicolumn{2}{c}{\makecell{\textbf{Random modality}\\\textbf{dropout}}} \\
\cmidrule(lr){4-5}\cmidrule(lr){6-7}\cmidrule(lr){8-9}\cmidrule(lr){10-11}
 &  &  &
\textbf{IoU} & \textbf{F$_1$} &
\textbf{IoU} & \textbf{F$_1$} &
\textbf{IoU} & \textbf{F$_1$} &
\textbf{IoU} & \textbf{F$_1$} \\
\midrule
\cmark & \cmark & \cmark &
\textbf{0.915 $\pm$ 0.231} & \textbf{0.930 $\pm$ 0.215} &
0.881 $\pm$ 0.289 & 0.894 $\pm$ 0.279 &
0.885 $\pm$ 0.287 & 0.896 $\pm$ 0.279 &
0.903 $\pm$ 0.255 & 0.916 $\pm$ 0.242 \\
\xmark & \cmark & \cmark &
0.508 $\pm$ 0.176 & 0.654 $\pm$ 0.169 &
0.880 $\pm$ 0.292 & 0.892 $\pm$ 0.284 &
0.525 $\pm$ 0.149 & 0.674 $\pm$ 0.145 &
\textbf{0.901 $\pm$ 0.256} & \textbf{0.915 $\pm$ 0.241} \\
\cmark & \xmark & \cmark &
0.847 $\pm$ 0.274 & 0.881 $\pm$ 0.245 &
0.871 $\pm$ 0.299 & 0.885 $\pm$ 0.289 &
0.780 $\pm$ 0.376 & 0.800 $\pm$ 0.366 &
\textbf{0.893 $\pm$ 0.269} & \textbf{0.907 $\pm$ 0.255} \\
\cmark & \cmark & \xmark &
0.882 $\pm$ 0.254 & 0.906 $\pm$ 0.234 &
0.881 $\pm$ 0.291 & 0.893 $\pm$ 0.282 &
0.855 $\pm$ 0.307 & 0.873 $\pm$ 0.295 &
\textbf{0.902 $\pm$ 0.256} & \textbf{0.916 $\pm$ 0.241} \\
\cmark & \xmark & \xmark &
0.680 $\pm$ 0.408 & 0.713 $\pm$ 0.398 &
0.872 $\pm$ 0.298 & 0.886 $\pm$ 0.287 &
0.702 $\pm$ 0.427 & 0.720 $\pm$ 0.419 &
\textbf{0.876 $\pm$ 0.300} & \textbf{0.887 $\pm$ 0.293} \\
\xmark & \cmark & \xmark &
0.522 $\pm$ 0.262 & 0.642 $\pm$ 0.254 &
0.879 $\pm$ 0.294 & 0.891 $\pm$ 0.286 &
0.543 $\pm$ 0.194 & 0.681 $\pm$ 0.184 &
\textbf{0.896 $\pm$ 0.265} & \textbf{0.910 $\pm$ 0.251} \\
\xmark & \xmark & \cmark &
0.473 $\pm$ 0.113 & 0.633 $\pm$ 0.125 &
0.856 $\pm$ 0.323 & 0.868 $\pm$ 0.313 &
0.561 $\pm$ 0.153 & 0.704 $\pm$ 0.149 &
\textbf{0.865 $\pm$ 0.306} & \textbf{0.879 $\pm$ 0.294} \\
\bottomrule
\end{tabular}

\vspace{1mm}
\footnotesize
\emph{Note:} No modality dropout and random modality dropout training settings are used.
The best scores under each modality availability combination are marked as bold.
\xmark\ and \cmark\ are used to represent missing and available modalities in the DSTL image set, respectively.

\end{table*}

\begin{table*}[t]
\centering
\caption{COMPARISON OF OUR PROPOSED MODEL WITH BASELINE ON POTSDAM TEST SET FOR TREE CLASS}
\label{tab:potsdam_tree_perf}
\setlength{\tabcolsep}{4.8pt}
\renewcommand{\arraystretch}{1.15}
\begin{tabular}{ccc cccc cccc}
\toprule
\multicolumn{3}{c}{\textbf{Modalities}} &
\multicolumn{4}{c}{\textbf{CBC (baseline)}} &
\multicolumn{4}{c}{\textbf{CBC-SLP (ours)}} \\
\cmidrule(lr){1-3}\cmidrule(lr){4-7}\cmidrule(lr){8-11}
\textbf{RGB} & \textbf{IRRG} & \textbf{DSM} &
\multicolumn{2}{c}{\makecell{\textbf{No modality}\\\textbf{dropout}}} &
\multicolumn{2}{c}{\makecell{\textbf{Random modality}\\\textbf{dropout}}} &
\multicolumn{2}{c}{\makecell{\textbf{No modality}\\\textbf{dropout}}} &
\multicolumn{2}{c}{\makecell{\textbf{Random modality}\\\textbf{dropout}}} \\
\cmidrule(lr){4-5}\cmidrule(lr){6-7}\cmidrule(lr){8-9}\cmidrule(lr){10-11}
 &  &  &
\textbf{IoU} & \textbf{F$_1$} &
\textbf{IoU} & \textbf{F$_1$} &
\textbf{IoU} & \textbf{F$_1$} &
\textbf{IoU} & \textbf{F$_1$} \\
\midrule
\cmark & \cmark & \cmark &
\textbf{0.802 $\pm$ 0.292} & \textbf{0.846 $\pm$ 0.276} &
0.762 $\pm$ 0.331 & 0.805 $\pm$ 0.320 &
0.671 $\pm$ 0.351 & 0.734 $\pm$ 0.336 &
0.766 $\pm$ 0.325 & 0.811 $\pm$ 0.312 \\
\xmark & \cmark & \cmark &
0.643 $\pm$ 0.375 & 0.701 $\pm$ 0.360 &
0.758 $\pm$ 0.332 & 0.802 $\pm$ 0.321 &
0.564 $\pm$ 0.334 & 0.653 $\pm$ 0.326 &
\textbf{0.759 $\pm$ 0.331} & \textbf{0.803 $\pm$ 0.317} \\
\cmark & \xmark & \cmark &
0.587 $\pm$ 0.390 & 0.649 $\pm$ 0.374 &
0.711 $\pm$ 0.361 & 0.758 $\pm$ 0.349 &
0.533 $\pm$ 0.374 & 0.608 $\pm$ 0.362 &
\textbf{0.713 $\pm$ 0.362} & \textbf{0.759 $\pm$ 0.351} \\
\cmark & \cmark & \xmark &
0.678 $\pm$ 0.359 & 0.737 $\pm$ 0.336 &
0.753 $\pm$ 0.339 & 0.796 $\pm$ 0.329 &
0.620 $\pm$ 0.358 & 0.691 $\pm$ 0.340 &
\textbf{0.761 $\pm$ 0.329} & \textbf{0.806 $\pm$ 0.316} \\
\cmark & \xmark & \xmark &
0.477 $\pm$ 0.456 & 0.508 $\pm$ 0.448 &
0.682 $\pm$ 0.386 & 0.725 $\pm$ 0.379 &
0.443 $\pm$ 0.354 & 0.527 $\pm$ 0.359 &
\textbf{0.691 $\pm$ 0.376} & \textbf{0.737 $\pm$ 0.366} \\
\xmark & \cmark & \xmark &
0.530 $\pm$ 0.434 & 0.574 $\pm$ 0.421 &
0.745 $\pm$ 0.343 & 0.789 $\pm$ 0.333 &
0.539 $\pm$ 0.380 & 0.611 $\pm$ 0.368 &
\textbf{0.750 $\pm$ 0.337} & \textbf{0.795 $\pm$ 0.324} \\
\xmark & \xmark & \cmark &
0.499 $\pm$ 0.430 & 0.546 $\pm$ 0.418 &
\textbf{0.584 $\pm$ 0.420} & \textbf{0.628 $\pm$ 0.415} &
0.445 $\pm$ 0.344 & 0.536 $\pm$ 0.340 &
0.576 $\pm$ 0.432 & 0.614 $\pm$ 0.427 \\
\bottomrule
\end{tabular}

\vspace{1mm}
\footnotesize
\emph{Note:} No modality dropout and random modality dropout training settings are used.
The best scores under each modality availability combination are marked as bold.
\xmark\ and \cmark\ are used to represent missing and available modalities in the Potsdam image set, respectively.
\end{table*}

\begin{table*}[t]
\centering
\caption{COMPARISON OF OUR PROPOSED MODEL WITH BASELINE ON HUNAN TEST SET FOR FOREST CLASS}
\label{tab:hunan_forest_perf}
\setlength{\tabcolsep}{4.8pt}
\renewcommand{\arraystretch}{1.15}
\begin{tabular}{ccc cccc cccc}
\toprule
\multicolumn{3}{c}{\textbf{Modalities}} &
\multicolumn{4}{c}{\textbf{CBC (baseline)}} &
\multicolumn{4}{c}{\textbf{CBC-SLP (ours)}} \\
\cmidrule(lr){1-3}\cmidrule(lr){4-7}\cmidrule(lr){8-11}
\textbf{SAR} & \textbf{MSI} & \textbf{DEM} &
\multicolumn{2}{c}{\makecell{\textbf{No modality}\\\textbf{dropout}}} &
\multicolumn{2}{c}{\makecell{\textbf{Random modality}\\\textbf{dropout}}} &
\multicolumn{2}{c}{\makecell{\textbf{No modality}\\\textbf{dropout}}} &
\multicolumn{2}{c}{\makecell{\textbf{Random modality}\\\textbf{dropout}}} \\
\cmidrule(lr){4-5}\cmidrule(lr){6-7}\cmidrule(lr){8-9}\cmidrule(lr){10-11}
 &  &  &
\textbf{IoU} & \textbf{F$_1$} &
\textbf{IoU} & \textbf{F$_1$} &
\textbf{IoU} & \textbf{F$_1$} &
\textbf{IoU} & \textbf{F$_1$} \\
\midrule
\cmark & \cmark & \cmark &
0.642 $\pm$ 0.278 & 0.738 $\pm$ 0.264 &
0.621 $\pm$ 0.309 & 0.709 $\pm$ 0.300 &
0.635 $\pm$ 0.272 & 0.735 $\pm$ 0.257 &
\textbf{0.659 $\pm$ 0.299} & \textbf{0.743 $\pm$ 0.287} \\
\xmark & \cmark & \cmark &
0.623 $\pm$ 0.291 & 0.718 $\pm$ 0.280 &
0.606 $\pm$ 0.314 & 0.696 $\pm$ 0.304 &
0.609 $\pm$ 0.281 & 0.710 $\pm$ 0.270 &
\textbf{0.640 $\pm$ 0.304} & \textbf{0.727 $\pm$ 0.295} \\
\cmark & \xmark & \cmark &
0.439 $\pm$ 0.314 & 0.546 $\pm$ 0.304 &
0.526 $\pm$ 0.319 & 0.623 $\pm$ 0.323 &
0.524 $\pm$ 0.297 & 0.631 $\pm$ 0.297 &
\textbf{0.582 $\pm$ 0.323} & \textbf{0.670 $\pm$ 0.326} \\
\cmark & \cmark & \xmark &
0.641 $\pm$ 0.267 & 0.739 $\pm$ 0.258 &
0.604 $\pm$ 0.321 & 0.690 $\pm$ 0.317 &
0.620 $\pm$ 0.263 & 0.724 $\pm$ 0.258 &
\textbf{0.652 $\pm$ 0.300} & \textbf{0.737 $\pm$ 0.291} \\
\cmark & \xmark & \xmark &
0.369 $\pm$ 0.269 & 0.488 $\pm$ 0.267 &
0.497 $\pm$ 0.315 & 0.599 $\pm$ 0.318 &
0.435 $\pm$ 0.246 & 0.563 $\pm$ 0.259 &
\textbf{0.569 $\pm$ 0.298} & \textbf{0.668 $\pm$ 0.302} \\
\xmark & \cmark & \xmark &
0.601 $\pm$ 0.292 & 0.697 $\pm$ 0.295 &
0.590 $\pm$ 0.324 & 0.679 $\pm$ 0.316 &
0.582 $\pm$ 0.280 & 0.686 $\pm$ 0.283 &
\textbf{0.616 $\pm$ 0.313} & \textbf{0.704 $\pm$ 0.306} \\
\xmark & \xmark & \cmark &
0.411 $\pm$ 0.321 & 0.511 $\pm$ 0.323 &
0.517 $\pm$ 0.345 & 0.603 $\pm$ 0.350 &
0.392 $\pm$ 0.237 & 0.519 $\pm$ 0.267 &
\textbf{0.518 $\pm$ 0.322} & \textbf{0.616 $\pm$ 0.321} \\
\bottomrule
\end{tabular}

\vspace{1mm}
\footnotesize
\emph{Note:} No modality dropout and random modality dropout training settings are used.
The best scores under each modality availability combination are marked as bold.
\xmark\ and \cmark\ are used to represent missing and available modalities in the Hunan image set, respectively.

\end{table*}

\subsection{Performance Comparisons With Baseline Model}

Table~\ref{tab:dstl_crop_perf} reports DSTL test set results for the baseline CBC model \cite{ulku2025cross} and the proposed CBC-SLP model. For each modality availability scenario, results are analyzed both with and without the random modality dropout training approach. Overall, CBC-SLP delivers the best performance across scenarios. These findings indicate that the proposed structured latent projection strategy is beneficial when training uses a random modality dropout approach.
By gating the projection on the modality availability mask in each batch, our model appears to transfer richer information to the decoder.

Training the baseline CBC model with a random modality dropout approach decreases IoU by 3.4\% when all modalities are available. This behavior is expected because CBC forwards only the shared representation to the decoder, creating a trade-off between the full modality and missing modality cases. In contrast, when training our CBC-SLP model with a random modality dropout approach, it achieves an IoU gain of 1.8\% under the full modality setting. Even when only the SWIR modality is available, our CBC-SLP still improves IoU by 0.9\% over the baseline. These results suggest that our model remains robust under all possible scenarios.

A similar performance pattern is observed in Table~\ref{tab:potsdam_tree_perf} and Table~\ref{tab:hunan_forest_perf} for Potsdam and Hunan test sets, respectively.
For the Potsdam image set, the CBC-SLP achieves 0.4\% better IoU than the baseline in full modality availability and performs better in almost all other scenarios. On the Hunan test set, our model outperforms the baseline in all evaluated scenarios, including an IoU gain of 3.8\% when all modalities are available.

\begin{table*}[t]
\centering
\caption{{DSTL TEST SET RESULTS} UNDER SEPARATE AND UNIFIED STRATEGIES}
\label{tab:dstl_ablation_separate_unified}
\setlength{\tabcolsep}{4.8pt}
\renewcommand{\arraystretch}{1.15}
\begin{tabular}{ccc cc cc cc}
\toprule
\multicolumn{3}{c}{\textbf{Modalities}} &
\multicolumn{2}{c}{\textbf{CBC (baseline)}} &
\multicolumn{2}{c}{\textbf{CBC-SLP (ours)}} &
\multicolumn{2}{c}{\textbf{CBC-SLP (ours)}} \\
\cmidrule(lr){1-3}\cmidrule(lr){4-5}\cmidrule(lr){6-7}\cmidrule(lr){8-9}
\textbf{RGB} & \textbf{NIR} & \textbf{SWIR} &
\multicolumn{2}{c}{\makecell{\textbf{Not unified}}} &
\multicolumn{2}{c}{\makecell{\textbf{Not unified}}} &
\multicolumn{2}{c}{\makecell{\textbf{Random modality dropout}}} \\
\cmidrule(lr){4-5}\cmidrule(lr){6-7}\cmidrule(lr){8-9}
 &  &  &
\textbf{IoU} & \textbf{F$_1$} &
\textbf{IoU} & \textbf{F$_1$} &
\textbf{IoU} & \textbf{F$_1$} \\
\midrule
\cmark & \cmark & \cmark &
\textbf{0.915 $\pm$ 0.231} & \textbf{0.930 $\pm$ 0.215} &
0.885 $\pm$ 0.287 & 0.896 $\pm$ 0.279 &
0.903 $\pm$ 0.255 & 0.916 $\pm$ 0.242 \\
\xmark & \cmark & \cmark &
0.824 $\pm$ 0.329 & 0.847 $\pm$ 0.316 &
0.676 $\pm$ 0.444 & 0.693 $\pm$ 0.430 &
\textbf{0.901 $\pm$ 0.256} & \textbf{0.915 $\pm$ 0.241} \\
\cmark & \xmark & \cmark &
0.658 $\pm$ 0.463 & 0.665 $\pm$ 0.461 &
0.660 $\pm$ 0.472 & 0.661 $\pm$ 0.471 &
\textbf{0.893 $\pm$ 0.269} & \textbf{0.907 $\pm$ 0.255} \\
\cmark & \cmark & \xmark &
0.674 $\pm$ 0.442 & 0.690 $\pm$ 0.437 &
0.853 $\pm$ 0.271 & 0.883 $\pm$ 0.260 &
\textbf{0.902 $\pm$ 0.256} & \textbf{0.916 $\pm$ 0.241} \\
\cmark & \xmark & \xmark &
0.661 $\pm$ 0.461 & 0.669 $\pm$ 0.458 &
0.623 $\pm$ 0.431 & 0.653 $\pm$ 0.432 &
\textbf{0.876 $\pm$ 0.300} & \textbf{0.887 $\pm$ 0.293} \\
\xmark & \cmark & \xmark &
0.619 $\pm$ 0.436 & 0.648 $\pm$ 0.433 &
0.511 $\pm$ 0.157 & 0.660 $\pm$ 0.159 &
\textbf{0.896 $\pm$ 0.265} & \textbf{0.910 $\pm$ 0.251} \\
\xmark & \xmark & \cmark &
0.824 $\pm$ 0.325 & 0.849 $\pm$ 0.310 &
0.764 $\pm$ 0.297 & 0.819 $\pm$ 0.283 &
\textbf{0.865 $\pm$ 0.306} & \textbf{0.879 $\pm$ 0.294} \\
\bottomrule
\end{tabular}

\vspace{1mm}
\footnotesize
\emph{Note:} The best scores under each modality availability combination are marked as bold.
\xmark\ and \cmark\ are used to represent missing and available modalities in the DSTL image set, respectively.
\end{table*}

\begin{table*}[t]
\centering
\caption{{POTSDAM TEST SET RESULTS} UNDER SEPARATE AND UNIFIED STRATEGIES}
\label{tab:potsdam_ablation_separate_unified}
\setlength{\tabcolsep}{4.8pt}
\renewcommand{\arraystretch}{1.15}
\begin{tabular}{ccc cc cc cc}
\toprule
\multicolumn{3}{c}{\textbf{Modalities}} &
\multicolumn{2}{c}{\textbf{CBC (baseline)}} &
\multicolumn{2}{c}{\textbf{CBC-SLP (ours)}} &
\multicolumn{2}{c}{\textbf{CBC-SLP (ours)}} \\
\cmidrule(lr){1-3}\cmidrule(lr){4-5}\cmidrule(lr){6-7}\cmidrule(lr){8-9}
\textbf{RGB} & \textbf{IRRG} & \textbf{DSM} &
\multicolumn{2}{c}{\makecell{\textbf{Not unified}}} &
\multicolumn{2}{c}{\makecell{\textbf{Not unified}}} &
\multicolumn{2}{c}{\makecell{\textbf{Random modality dropout}}} \\
\cmidrule(lr){4-5}\cmidrule(lr){6-7}\cmidrule(lr){8-9}
 &  &  &
\textbf{IoU} & \textbf{F$_1$} &
\textbf{IoU} & \textbf{F$_1$} &
\textbf{IoU} & \textbf{F$_1$} \\
\midrule
\cmark & \cmark & \cmark &
\textbf{0.802 $\pm$ 0.292} & \textbf{0.846 $\pm$ 0.276} &
0.671 $\pm$ 0.351 & 0.734 $\pm$ 0.336 &
0.766 $\pm$ 0.325 & 0.811 $\pm$ 0.312 \\
\xmark & \cmark & \cmark &
0.627 $\pm$ 0.374 & 0.692 $\pm$ 0.349 &
0.684 $\pm$ 0.305 & 0.761 $\pm$ 0.287 &
\textbf{0.759 $\pm$ 0.331} & \textbf{0.803 $\pm$ 0.317} \\
\cmark & \xmark & \cmark &
\textbf{0.747 $\pm$ 0.318} & \textbf{0.801 $\pm$ 0.298} &
0.636 $\pm$ 0.330 & 0.717 $\pm$ 0.307 &
0.713 $\pm$ 0.362 & 0.759 $\pm$ 0.351 \\
\cmark & \cmark & \xmark &
0.744 $\pm$ 0.331 & 0.795 $\pm$ 0.313 &
\textbf{0.805 $\pm$ 0.291} & \textbf{0.848 $\pm$ 0.274} &
0.761 $\pm$ 0.329 & 0.806 $\pm$ 0.316 \\
\cmark & \xmark & \xmark &
\textbf{0.750 $\pm$ 0.335} & \textbf{0.796 $\pm$ 0.320} &
0.701 $\pm$ 0.367 & 0.748 $\pm$ 0.357 &
0.691 $\pm$ 0.376 & 0.737 $\pm$ 0.366 \\
\xmark & \cmark & \xmark &
0.485 $\pm$ 0.458 & 0.518 $\pm$ 0.442 &
\textbf{0.767 $\pm$ 0.309} & \textbf{0.818 $\pm$ 0.291} &
0.750 $\pm$ 0.337 & 0.795 $\pm$ 0.324 \\
\xmark & \xmark & \cmark &
\textbf{0.591 $\pm$ 0.415} & \textbf{0.637 $\pm$ 0.408} &
0.514 $\pm$ 0.308 & 0.616 $\pm$ 0.317 &
0.576 $\pm$ 0.432 & 0.614 $\pm$ 0.427 \\
\bottomrule
\end{tabular}

\vspace{1mm}
\footnotesize
\emph{Note:} The best scores under each modality availability combination are marked as bold.
\xmark\ and \cmark\ are used to represent missing and available modalities in the Potsdam image set, respectively.
\end{table*}

\begin{table*}[t]
\centering
\caption{{HUNAN TEST SET RESULTS} UNDER SEPARATE AND UNIFIED STRATEGIES}
\label{tab:hunan_ablation_separate_unified}
\setlength{\tabcolsep}{4.8pt}
\renewcommand{\arraystretch}{1.15}
\begin{tabular}{ccc cc cc cc}
\toprule
\multicolumn{3}{c}{\textbf{Modalities}} &
\multicolumn{2}{c}{\textbf{CBC (baseline)}} &
\multicolumn{2}{c}{\textbf{CBC-SLP (ours)}} &
\multicolumn{2}{c}{\textbf{CBC-SLP (ours)}} \\
\cmidrule(lr){1-3}\cmidrule(lr){4-5}\cmidrule(lr){6-7}\cmidrule(lr){8-9}
\textbf{SAR} & \textbf{MSI} & \textbf{DEM} &
\multicolumn{2}{c}{\makecell{\textbf{Not unified}}} &
\multicolumn{2}{c}{\makecell{\textbf{Not unified}}} &
\multicolumn{2}{c}{\makecell{\textbf{Random modality dropout}}} \\
\cmidrule(lr){4-5}\cmidrule(lr){6-7}\cmidrule(lr){8-9}
 &  &  &
\textbf{IoU} & \textbf{F$_1$} &
\textbf{IoU} & \textbf{F$_1$} &
\textbf{IoU} & \textbf{F$_1$} \\
\midrule
\cmark & \cmark & \cmark &
0.642 $\pm$ 0.278 & 0.738 $\pm$ 0.264 &
0.635 $\pm$ 0.272 & 0.735 $\pm$ 0.257 &
\textbf{0.659 $\pm$ 0.299} & \textbf{0.743 $\pm$ 0.287} \\
\xmark & \cmark & \cmark &
\textbf{0.662 $\pm$ 0.275} & \textbf{0.752 $\pm$ 0.271} &
0.637 $\pm$ 0.272 & 0.733 $\pm$ 0.271 &
0.640 $\pm$ 0.304 & 0.727 $\pm$ 0.295 \\
\cmark & \xmark & \cmark &
\textbf{0.618 $\pm$ 0.291} & \textbf{0.713 $\pm$ 0.285} &
0.558 $\pm$ 0.304 & 0.657 $\pm$ 0.307 &
0.582 $\pm$ 0.323 & 0.670 $\pm$ 0.326 \\
\cmark & \cmark & \xmark &
0.625 $\pm$ 0.291 & 0.719 $\pm$ 0.283 &
\textbf{0.660 $\pm$ 0.280} & \textbf{0.750 $\pm$ 0.270} &
0.652 $\pm$ 0.300 & 0.737 $\pm$ 0.291 \\
\cmark & \xmark & \xmark &
0.423 $\pm$ 0.333 & 0.520 $\pm$ 0.325 &
0.560 $\pm$ 0.279 & 0.668 $\pm$ 0.287 &
\textbf{0.569 $\pm$ 0.298} & \textbf{0.668 $\pm$ 0.302} \\
\xmark & \cmark & \xmark &
0.514 $\pm$ 0.302 & 0.619 $\pm$ 0.304 &
0.578 $\pm$ 0.309 & 0.674 $\pm$ 0.304 &
\textbf{0.616 $\pm$ 0.313} & \textbf{0.704 $\pm$ 0.306} \\
\xmark & \xmark & \cmark &
\textbf{0.545 $\pm$ 0.32} & \textbf{0.638 $\pm$ 0.327} &
0.476 $\pm$ 0.269 & 0.594 $\pm$ 0.287 &
0.518 $\pm$ 0.322 & 0.616 $\pm$ 0.321 \\
\bottomrule
\end{tabular}

\vspace{1mm}
\footnotesize
\emph{Note:} The best scores under each modality availability combination are marked as bold.
\xmark\ and \cmark\ are used to represent missing and available modalities in the Hunan image set, respectively.

\end{table*}

\subsection{Separate vs Unified Models}

{In a separate experiment, we analyze whether training a separate model for each modality availability scenario yields measurable gains over training a single model with random modality dropout.} Intuitively, training unique models for each scenario separately might be expected to define an upper bound for each case. However, the results in Table~\ref{tab:dstl_ablation_separate_unified}, Table~\ref{tab:potsdam_ablation_separate_unified}, and Table~\ref{tab:hunan_ablation_separate_unified} for DSTL, Potsdam, and Hunan image sets, respectively,  show that this expectation does not hold consistently across scenarios.

On the DSTL test set, our CBC-SLP model reaches the best performance in all missing modality scenarios, except the full modality case. For the Potsdam image set, models specific to scenarios often provide the numerical upper bound, but the performance gap with our single CBC-SLP model remains limited. For instance, in the scenario where only the DSM modality is available, the IoU difference between our unified model and the base model trained specifically for that scenario is only 1.5\%. From an efficiency perspective, this margin can be considered small relative to training multiple models specific to each scenario. On the Hunan image set, our model sets the upper bounds in the full modality case and in the cases where either only SAR or MSI is available. In the remaining scenarios, the performance differences can be considered negligible when compared with training a separate model for each modality combination. Therefore, these results support the same conclusion across all three image sets. A single CBC-SLP model trained with random modality dropout can provide a robust trade-off between accuracy and efficiency.

\subsection{Information Gap Analysis Under Different Modality Availability Scenarios}

In Fig.~\ref{fig:gap}, we measure how much information is present in the decoder input representation under each modality availability scenario $m$. Instead of using only the final predicted mask, we analyze the internal decoder representation because this is exactly where the base model CBC and our CBC-SLP model differ structurally. CBC-SLP forwards shared and modality-specific information, while CBC forwards only shared information. 

For each scenario $m$, we extract features from both models and train a $1\times1$ convolution to predict the mask from those features. The prediction error of this auxiliary classifier, measured with BCE, is used as a surrogate for conditional entropy $H(\mathbf{Y}\mid \mathbf{U}_m)$. Here, $H(\mathbf{Y}\mid \mathbf{U}_m)$ is the conditional entropy of target $\mathbf{Y}$ given the $\mathbf{U}_m$ as input. For one sample $i$, BCE is computed pixelwise and averaged over all spatial positions:
\begin{equation}
\begin{aligned}
\mathrm{BCE}_i
&=
-\frac{1}{|\Omega_i|}
\sum_{u\in\Omega_i}
\Big[
\mathbf{Y}_i(u)\log \hat{P}_i(u) \\
&\qquad\qquad
+\big(1-\mathbf{Y}_i(u)\big)\log\!\big(1-\hat{P}_i(u)\big)
\Big].
\end{aligned}
\label{eq:bce_pixelwise}
\end{equation}
where $\Omega_i$ is the set of pixel locations of sample $i$, $|\Omega_i|$ is the number of pixels in that sample, and $\hat{P}_i(u)$ is the auxiliary classifier's predicted probability for the positive class at pixel $u$. Then, the average over test samples for scenario $m$ is computed as:
\begin{equation}
\widehat{H}(\mathbf{Y}\mid \mathbf{U}_m)\approx \frac{1}{N_m}\sum_{i=1}^{N_m}\mathrm{BCE}_i.
\end{equation}
where $N_m$ is the total number of samples for the scenario $m$. Finally, the information gap is defined as follows:
\begin{equation}
\Delta_{\mathrm{p}}(m)
=
\widehat{H}(\mathbf{Y}\mid \mathbf{U}^{\mathrm{CBC}}_m)
-
\widehat{H}(\mathbf{Y}\mid \mathbf{U}^{\mathrm{CBC\_SLP}}_m).
\end{equation}

If $\Delta_{\mathrm{p}}(m)>0$, CBC\_SLP has lower conditional uncertainty at the decoder representation for scenario $m$, meaning that the representation is more informative about target $\mathbf{Y}$.

In implementation, we resize the mask to the feature resolution using nearest-neighbor interpolation before computing BCE, ensuring that the loss is always computed on aligned tensors.

As shown in Fig.~\ref{fig:gap}, the calculated information gap is generally positive across modality availability scenarios. This common trend suggests that the decoder representation of CBC-SLP is semantically more informative about the target than that of CBC in most cases (as can be seen in Fig.~\ref{fig:gap_dstl} and Fig.~\ref{fig:gap_hunan}). However, in the Potsdam image set (Fig.~\ref{fig:gap_potsdam}), the information gap is not positive in scenarios where mostly the RGB information is available. This RGB modality may contain limited discriminative information about the chosen target, and to some extent, its private branch seems to introduce nuisance variation rather than useful evidence. In these cases, we interpret that the shared features tend to address the discriminative parts, which dominate the prediction of the target.

\subsection{Qualitative Results}

Examples of qualitative results are depicted in Fig.~\ref{fig:dstl_visual}, Fig.~\ref{fig:potsdam_visual}, and Fig.~\ref{fig:hunan_visual} for DSTL, Hunan, and Potsdam image sets, respectively. It is possible to see the overall improvement in the predictions of our CBC-SLP model compared to those of the baseline in both full and missing modality scenarios. As shown in Fig.~\ref{fig:dstl_visual}, we observe that the CBC-SLP model can reduce incorrect and missed detections in different scenarios, whereas the others tend to produce noisier predictions for crop regions. Visualizations in Fig.~\ref{fig:potsdam_visual} demonstrate that CBC-SLP is able to recover tree boundaries with better precision at different modality availability scenarios according to the ground truth, while also yielding denser predicted target pixels. It is also worth noting that the tree region predictions yielded by our model in the first row seem to be much more clearly separated from similar non-tree surfaces than those produced by the baseline. In Fig.~\ref{fig:hunan_visual}, although predictions look quite noisy, our model can preserve fine spatial details of the forest structures more effectively than the compared models. Shapes are predicted accurately, both for large and small forest regions, even in challenging scenarios when only one or two modalities are available.

\section{Conclusion}
We propose a novel multimodal semantic segmentation strategy to maintain robust performance for both full and missing modality scenarios in remote sensing imagery. Existing models rely on learning shared representations. However, they may exhibit a performance trade-off when all modalities are available, as relying only on shared components does not account for modality-specific complementary information. Inspired by the theoretical findings on modality alignment, we propose a novel structured latent projection mechanism as an architectural inductive bias to transfer both shared and modality-specific information to the decoder according to the random modality availability mask. Our strategy ensures that the complementary information can be dynamically adapted without sacrificing the shared representations. We validated the proposed network design on three multimodal remote sensing image sets across all possible modality availability scenarios. Our CBC-SLP model outperforms state-of-the-art multimodal approaches by maintaining robustness when all modalities are available, rather than being tailored only for cases where modalities are missing. At the feature level, we also show that the proposed design can recover complementary information that may not be preserved when learning only a shared representation, as reflected by the information gap analysis in our experiments. Future research could focus on how the proposed structured latent projection approach performs in more diverse multimodal cases, such as integrating remote sensing imagery with text-based metadata.

\section*{Acknowledgment}
This study was supported by the Scientific and Technological Research Council of Turkey (TUBITAK) under the Grant Number 124E725. The authors thank TUBITAK for their support.

\bibliographystyle{IEEEtran}

\bibliography{Mybib}

\newpage

 

\begin{IEEEbiography}[{\includegraphics[width=1.1in,height=1.1in,clip,keepaspectratio]{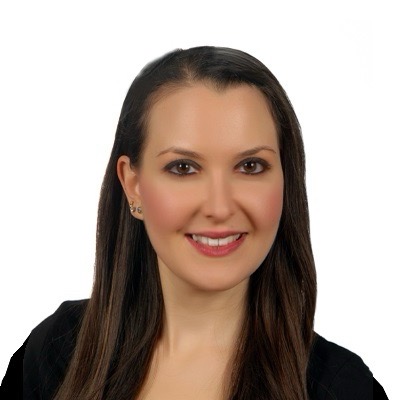}}]{{Irem Ulku}} 
received B.Sc. degrees in both Electronics and Communication Engineering and in Industrial Engineering from Çankaya University (Ankara, Turkey) in 2009 and 2010, respectively, followed by an M.Sc. degree in Electrical and Electronics Engineering from the Middle East Technical University (Ankara, Turkey) in 2013. She then received her PhD degree in Electronics and Communication Engineering from Çankaya University in 2017. In 2019, she was a Research Associate at Imperial College London, United Kingdom. She is currently an Assistant Professor at the Department of Computer Engineering, Ankara University (Turkey). Her research interests include hyperspectral image processing and deep learning-based semantic segmentation.
\end{IEEEbiography}

\begin{IEEEbiography}[{\includegraphics[width=1.1in,height=1.1in,clip,keepaspectratio]{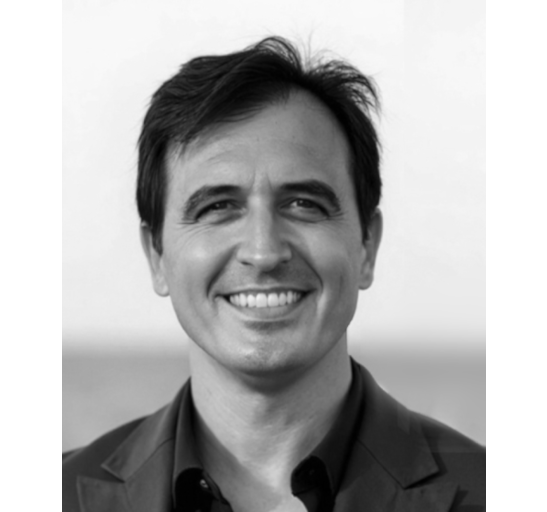}}]{{Erdem Akagündüz}} 
received his B.Sc., M.Sc., and Ph.D. degrees in Electrical and Electronics Engineering from Middle East Technical University in 2001, 2004, and 2010, respectively. Between 2001–2008, he worked as a research assistant at METU Computer Vision and Intelligent Systems Laboratory; between 2008–2009, he was a research fellow at the University of York, UK; and from 2009–2016, he served as an embedded algorithm development engineer at ASELSAN Inc. In 2016, he worked as a postdoctoral researcher at the University of York again. After serving at Çankaya University between 2018–2021, he joined METU Informatics Institute, where he currently works. His research interests include deep learning applications and embedded deep learning. Dr. Akagündüz holds several international patents and publications in these areas. 
\end{IEEEbiography}

\begin{IEEEbiography}[{\includegraphics[width=1.1in,height=1.1in,clip,keepaspectratio]{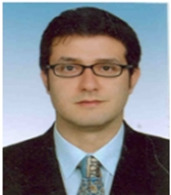}}]{{Ömer Özgür Tanrıöver}} 
received the B.Sc. degree in Computer Engineering, the M.Sc. and Ph.D. degrees in Information Systems from Middle East Technical University, Ankara, Turkey. Earlier, he was as a Certified Information Systems Auditor (CISA) with the Information Management Department, Banking Regulation Agency of Turkey. He is currently an Associate Professor in the Computer Engineering Department of Ankara University. His current research interests include applications of deep  learning in medical informatics, remote sensing, human computer interaction, and information system security.
\end{IEEEbiography}

\vfill

\end{document}